\documentclass[]{mamoda}

\usepackage{amssymb}
\usepackage{multirow}
\usepackage{bigdelim}
\usepackage{todonotes}
\usepackage{longtable}
\usepackage{tabularx}
\usepackage{wrapfig}
\usepackage[most]{tcolorbox} 
\usepackage{xcolor}
\usepackage{url}

\usepackage{tikz}
\usepackage{forest}

\usepackage{arydshln}
\usepackage{colortbl}
\usepackage{amssymb}
\usepackage{pifont}
\usepackage{booktabs,multirow}
\usepackage{caption} % for \captionof (used in side-by-side tables)
\usepackage{makecell}
\usepackage{tabulary}
\usepackage{fontawesome5}
\usepackage{bbding}
\usepackage{multicol}

\definecolor{prompt}{HTML}{5f84e4}
\definecolor{img}{HTML}{820100}
\definecolor{highlight}{HTML}{42B883}

\definecolor{CQColor}{rgb}{0.0,0.0,1.0}

\definecolor{TSColor}{rgb}{0.5,0.0,0.8}

\definecolor{CQRColor}{rgb}{1.0,0.0,1.0}

\definecolor{darkgreen}{RGB}{0, 150, 0} 

\newcommand \blfootnote[1]{
    \begingroup
        \renewcommand
        \thefootnote{}\footnote{#1}
        \addtocounter{footnote}{-1}
        \vspace{-1ex}
    \endgroup
}

\newlength\savewidth

\newcommand{\tablestyle}[2]{\setlength{\tabcolsep}{#1}\renewcommand{\arraystretch}{#2}\centering\footnotesize}

\title{MammothModa2: A Unified AR--Diffusion \\[0.15em] Framework for Multimodal Understanding and Generation}

\author[1]{MammothModa Team}
\affiliation[1]{ByteDance}

\abstract{Unified multimodal models aim to integrate understanding and generation within a single framework, yet bridging the gap between discrete semantic reasoning and high-fidelity visual synthesis remains challenging. We present MammothModa2 (Mammoth2), a unified autoregressive--diffusion (AR--Diffusion) framework designed to effectively couple autoregressive semantic planning with diffusion-based generation. Mammoth2 adopts a serial design: an AR path equipped with generation experts performs global semantic modeling over discrete tokens, while a single-stream Diffusion Transformer (DiT) decoder handles high-fidelity image synthesis. A carefully designed AR--Diffusion feature alignment module combines multi-layer feature aggregation, unified condition encoding, and in-context conditioning to stably align AR's representations with the diffusion decoder's continuous latents.\par
Mammoth2 is trained end-to-end with joint Next-Token Prediction and Flow Matching objectives, followed by supervised fine-tuning and reinforcement learning over both generation and editing. With roughly 60M supervised generation samples and no reliance on pre-trained generators, Mammoth2 delivers strong text-to-image and instruction-based editing performance on public benchmarks, achieving 0.87 on GenEval, 87.2 on DPGBench, and 4.06 on ImgEdit, while remaining competitive with understanding-only MLLM (e.g., Qwen3-VL-8B) on multimodal understanding tasks. These results suggest that a carefully coupled AR--Diffusion architecture can provide high-fidelity generation and editing while maintaining strong multimodal comprehension within a single, parameter- and data-efficient model.}

\checkdata[Code]{\url{https://github.com/bytedance/mammothmoda}}

\checkdata[HuggingFace]{\url{https://huggingface.co/bytedance-research/MammothModa}}
\begin{document}
\pagestyle{fancy}
\fancyhead{}
\fancyhead[R]{\footnotesize\color{black} Technical Report}
\maketitle
\renewcommand{\thefootnote}{}

\renewcommand{\thefootnote}{\arabic{footnote}}

% Introduction starts on the first page (below title/abstract)
\section{Introduction}
\label{sec:intro}

\begin{figure*}[!t]
    \centering
    \includegraphics[width=0.82\textwidth]{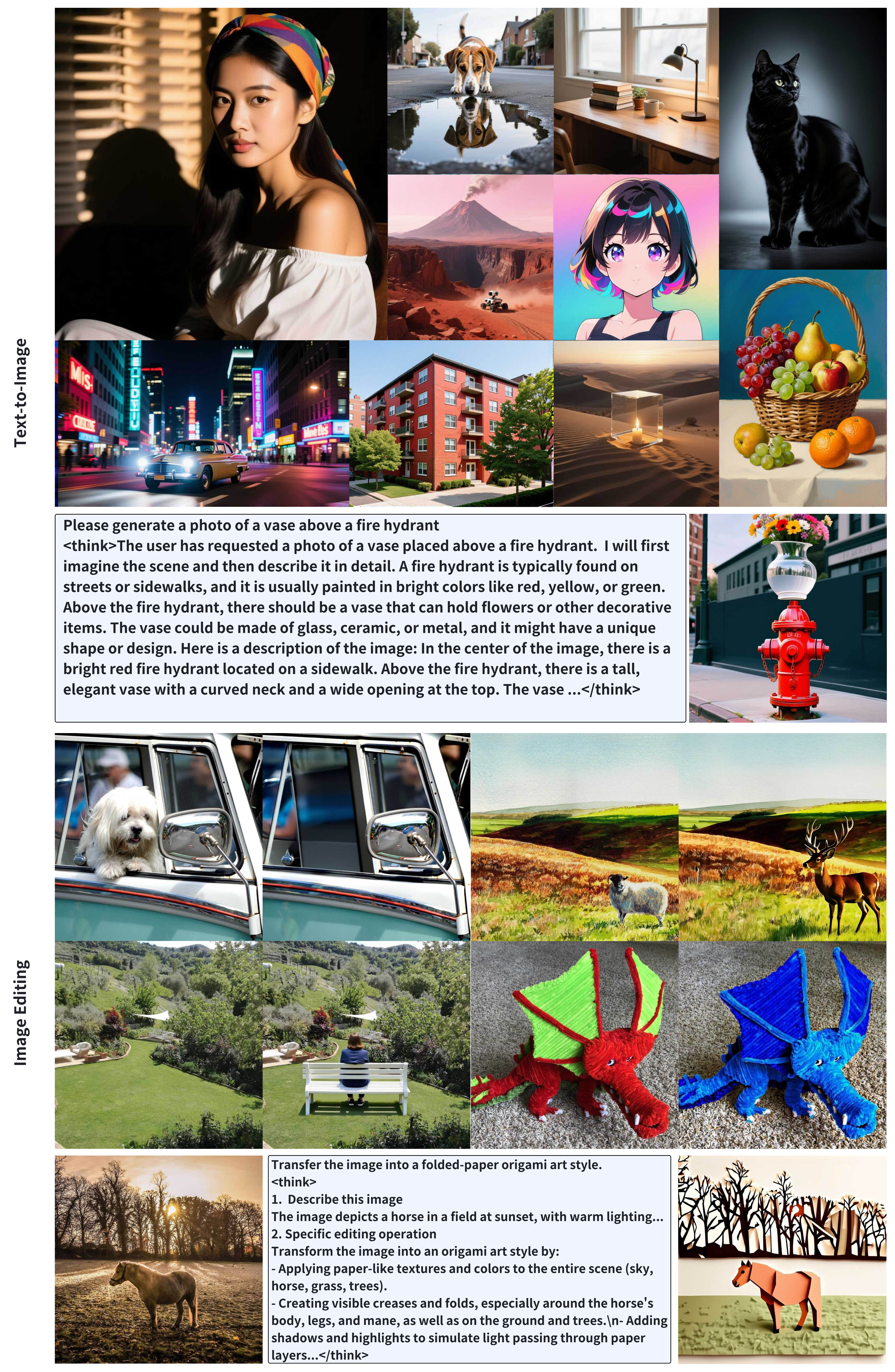}
    \caption{Mammoth2 seamlessly handles text-to-image generation, instruction-based editing, and visual understanding within a single model, delivering high-fidelity results across diverse real-world scenarios.}
    \vspace{-2mm}
    \label{fig:main_showcase}
\end{figure*}

Unified multimodal large language models (MLLMs) are rapidly converging into unified systems that integrate image understanding, generation, and editing within a single framework, enabling consistent semantic planning and execution under long-context instructions \cite{zhang2025unified}. Compared with task-specific pipelines that isolate capabilities and hinder knowledge transfer across modalities, unified models facilitate deep cross-modal collaboration and adapt more effectively to complex multimodal task requirements. As a result, designing a unified system that encompasses the complete ``understanding--planning--generation'' loop is widely regarded as a key step toward general multimodal intelligence.

Building upon vision-language understanding backbones through modality alignment and autoregressive (AR) modeling has become the de facto paradigm for multimodal understanding. Extending a vision-understanding foundation model to incorporate generative capability has emerged as a practical technical route for constructing a unified model \cite{lu2024ovis, wu2025omnigen2, lin2025uniworld}. Autoregressive and diffusion models represent two predominant generative paradigms. AR generation, driven by next-token prediction (NTP), naturally aligns with text LLMs and understanding architectures, providing strong global semantic planning and compositional control, but it suffers from error accumulation in long-sequence generation. In contrast, diffusion models---particularly Diffusion Transformers (DiTs) \cite{Peebles_2023_ICCV}---offer substantial advantages in fine-grained local details, texture quality, and high-resolution image synthesis, yet they are relatively weaker in complex instruction following and precise semantic control. Given their complementary strengths and limitations~\cite{feng2024ranni, brack2023segainstructingtexttoimagemodels}, integrating AR and diffusion has become a promising framework for unified vision-language generation, but it faces two central challenges: (i) how to inject generative pathways \emph{without degrading the backbone's multimodal understanding performance}, and (ii) how to achieve stable and consistent feature alignment between the discrete tokens in AR models (e.g., LLaMA \cite{touvron2023llama} sequences) and the continuous visual latents in diffusion models (e.g., SD-VAE \cite{rombach2021highresolution} latents), a difficulty also highlighted in prior work \cite{shi2025lmfusionadaptingpretrainedlanguage,li2025unifusionvisionlanguagemodelunified}.

We introduce \textbf{MammothModa2 (Mammoth2)}, a unified framework that integrates autoregressive (AR) and diffusion modeling through end-to-end joint training with NTP and Flow Matching \cite{lipman2022flow}. Mammoth2 serves as a general AR–Diffusion architecture that can be instantiated on diverse vision–language backbones; in this work, we build on the strong Qwen3-VL-8B \cite{qwen3blog2025} MLLM. We extend an understanding-centric backbone into a generator by introducing \textbf{generation experts} that process unified discrete visual tokens, enabling AR-based image synthesis while preserving the original \textbf{understanding experts} for perception and reasoning. The semantic hidden states produced by the AR pathway serve as conditioning signals for a DiT to perform high-fidelity pixel synthesis. To ensure effective feature alignment, we further design an AR–Diffusion alignment module that aggregates intermediate MLLM representations, unifies text–visual encodings, and injects them into the DiT, achieving deep cross-pathway fusion between AR and diffusion.

To further address the above challenges, we design a progressive, regularized joint training strategy and pair it with targeted data curation. Our training corpus is built around roughly 60M generation samples, complemented by additional multimodal understanding data. The overall recipe consists of two major stages: pre-training and post-training. The pre-training stage focuses on establishing strong generative capabilities atop the vision-understanding backbone. We adopt progressive learning strategies and introduce regularization for AR-based visual generation to alleviate the intrinsic train–inference mismatch that arises when AR and diffusion models are jointly trained, thereby stabilizing the shared feature space between the AR pathway and the diffusion decoder. The post-training stage further integrates multimodal understanding data and filtered high-quality generation samples for supervised fine-tuning. Finally, we apply reinforcement learning with a unified DiffusionNFT \cite{zheng2025diffusionnft} objective over both generation and editing tasks, leading to substantial performance gains in both capabilities.

Within this unified AR--Diffusion framework, Mammoth2 achieves state-of-the-art performance across understanding, generation, and editing. It reaches approximately 87\% on GenEval \cite{ghosh2024geneval} (0.87 overall), 87.2\% on DPGBench \cite{hu2024ella} (87.20 overall), and 4.06 on ImgEdit \cite{ye2025imgedit} benchmarks. Notably, Mammoth2 demonstrates strong parameter and data efficiency: it uses only about 60M supervised generation samples, does not rely on any pre-trained generative models, and delivers best-in-class results among models in a similar parameter range.

\vspace{2mm}
\noindent\textbf{Experimental Highlights.}
\begin{itemize}[leftmargin=1.5em, itemsep=0.25em, topsep=0.25em]
    \item \textbf{Unified AR--Diffusion Architecture with Expert Decoupling.} Mammoth2 couples an autoregressive backbone with a single-stream DiT decoder via the AR--Diffusion Feature Alignment Module, cleanly separating semantic planning from high-fidelity pixel synthesis within a shared discrete token space. Within the AR pathway, dedicated generation experts specialize in discrete visual token modeling while the original parameters remain focused on multimodal understanding, enabling strong generative capability without degrading perception and reasoning performance.
    
    \item \textbf{Joint Optimization and RL Alignment.} A multi-stage training recipe jointly optimizes Next-Token Prediction and Flow Matching objectives, followed by supervised fine-tuning and DiffusionNFT-based reinforcement learning on both generation and editing, yielding stable AR--Diffusion coordination and improved instruction adherence.
    
    \item \textbf{Strong Performance with Data and Parameter Efficiency.} Mammoth2 achieves 87\% on GenEval and 87.2\% on DPGBench for text-to-image generation, and 4.06 on ImgEdit, outperforming prior unified models such as BAGEL and OmniGen2 while approaching specialized proprietary systems on editing quality. These results are obtained with only 60M training samples and no pre-trained generator, demonstrating that Mammoth2 delivers strong unified multimodal performance under a practical, data- and parameter-efficient budget.
\end{itemize}
\section{Architecture}
\label{sec:method}

\subsection{Overview}

\begin{figure}[H]
\centering
\includegraphics[width=0.95\linewidth]{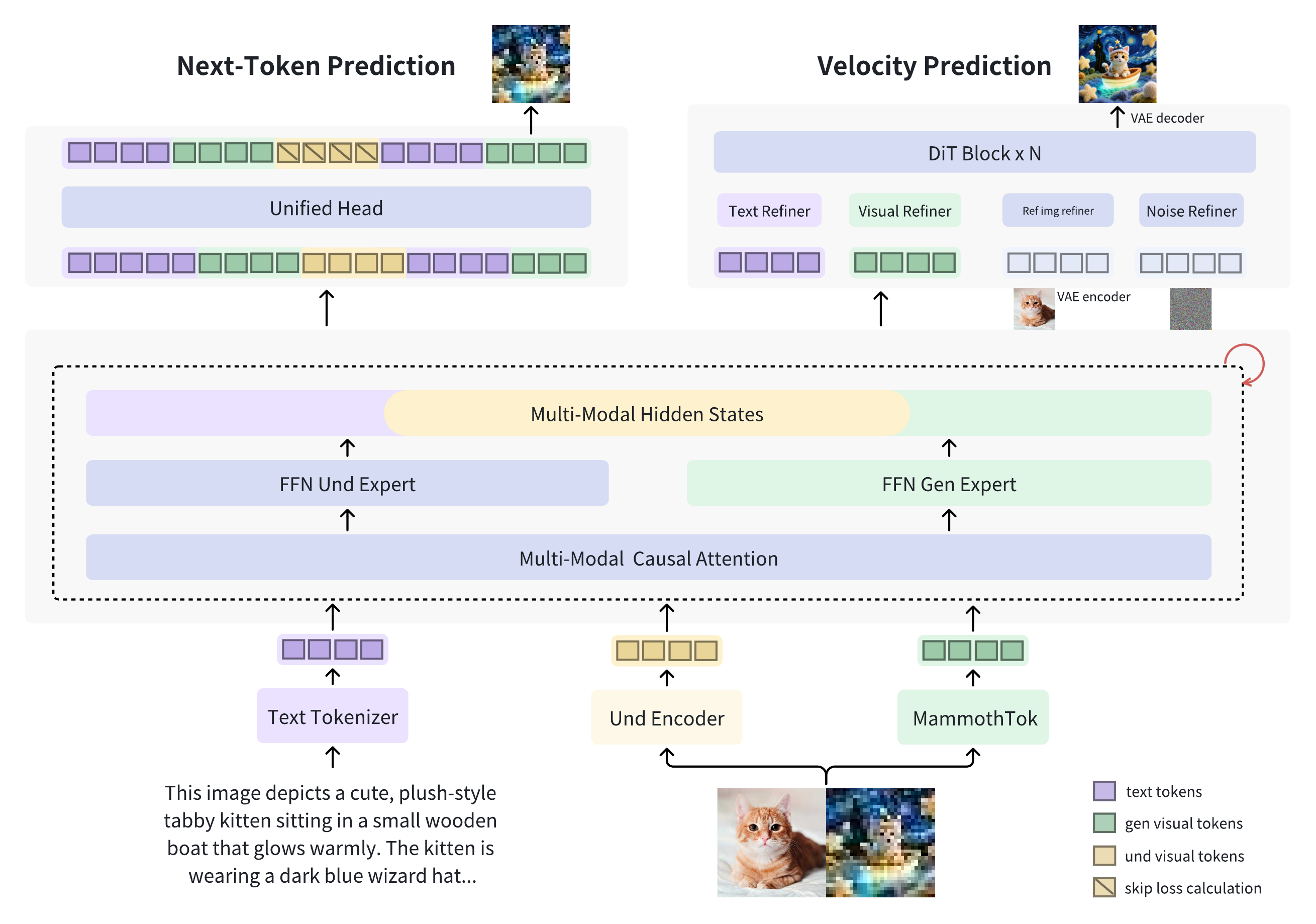}
\caption{\textbf{Mammoth2 Architecture.} A serial AR--Diffusion framework in which an autoregressive backbone performs semantic planning over MammothTok visual tokens, and a diffusion decoder generates high-fidelity images conditioned on the AR features.}
\label{fig:arch}
\end{figure}

Mammoth2 adopts a serial AR–diffusion architecture that supports both strong multimodal understanding and high-fidelity generation, as illustrated in Fig. \ref{fig:arch}. The system consists of: (1) Autoregressive Modeling component (Section~\ref{sec:architecture_ar}), which includes a Qwen3-VL-8B vision–language backbone augmented with a generative expert and a unified visual tokenizer (MammothTok) for discrete visual token autoregressive modeling; (2) Diffusion Generator (Section~\ref{sec:architecture_df}), implemented with a single-stream DiT architecture that produces high-precision images conditioned on the outputs of the MLLM's autoregressive modeling; and (3) AR–Diffusion Feature Alignment Module (Section~\ref{sec:architecture_ardf}), a carefully designed three-stage alignment architecture that serves as the key bridge between autoregressive modeling and diffusion generation. Overall, Mammoth2 requires only an additional 5B generative parameters ($\sim$3B from the AR generative expert and $\sim$2B from the diffusion decoder), providing a compact and practical blueprint for unified understanding, generation, and editing.

\subsection{Autoregressive Modeling}
\label{sec:architecture_ar}

\textbf{Vision-Language Model Backbone (VLM).} In this work, MammothModa2 (Mammoth2) is instantiated on Qwen3-VL-8B as its multimodal large language model backbone. This model offers enhanced visual understanding capabilities, including spatio-temporal perception across scales, precise object localization via bounding boxes and points, and strong text recognition. These capabilities provide a solid foundation for subsequent semantic planning and generation. More broadly, the Mammoth2 framework is agnostic to the particular choice of VLM backbone and can in principle be built atop other strong multimodal LLMs.

\textbf{Unified Visual Tokenizer.} Mammoth2 employs a discrete visual tokenizer as the image encoding scheme for generation, aiming to build a \textbf{unified tokenizer for both understanding and generation} that supports semantic alignment and high-quality reconstruction within a shared token space. Unlike traditional tokenizers that primarily focus on semantic alignment (\textit{e.g.}, SigLIP-VQ \cite{chen2025janus}), MammothTok is jointly optimized for native-resolution reconstruction and text rendering, thereby strengthening both reconstruction and generation ability. MammothTok uses a pre-trained AIMv2~\cite{fini2025multimodal} model as its encoder, supporting native-resolution inputs, and a hybrid CNN–ViT architecture as its decoder. Beyond standard VQ-GAN losses \cite{esser2021taming}, MammothTok incorporates a \textbf{semantic alignment loss} that injects semantic information from a pre-trained MLLM into discrete visual tokens, enhancing the tokenizer's semantic representation capacity. Detailed architecture configurations, loss definitions, training procedures, and experimental results for MammothTok are provided in Appendix~A.

\textbf{Generation Expert Architecture.} A key challenge in extending an understanding-centric MLLM to a unified generator is to avoid catastrophic forgetting of its perception and reasoning abilities while keeping parameter and compute overhead modest. To this end, following the Mixture-of-Experts (MoE) \cite{shazeer2017outrageously} paradigm, we duplicate the feed-forward networks (FFNs) in Transformer layers to instantiate dedicated generation experts. We adopt a \textbf{hard routing mechanism} \cite{deng2025emerging}: newly initialized generation experts are dedicated to processing discrete visual tokens from MammothTok, while the original parameters act as \textbf{understanding experts} that continue to handle image and text tokens for conventional multimodal understanding tasks. During autoregressive image generation, the model generates discrete tokens sequentially in raster-scan order (from top-left to bottom-right), and the tokenizer decoder reconstructs the predicted token sequence into a complete image. We optimize the autoregressive module with a standard next-token prediction (NTP) loss. Generation experts are deployed only in deeper layers \cite{li2025unifork}, which substantially reduces additional parameters and computational overhead while maintaining strong generation capability and preserving the shared early-layer representations used by understanding tasks.

\subsection{Diffusion Generator}
\label{sec:architecture_df}

Mammoth2's diffusion decoder adopts a single-stream architecture \cite{qin2025lumina}, where processed conditioning signals (features from the MLLM backbone) and noisy latent representations (VAE-encoded latents) are handled within the same sequence. Specifically, conditioning tokens and noisy latent patches are concatenated to form a unified input sequence, and the DiT performs \textbf{full-sequence self-attention} over the entire sequence, allowing conditioning information and generative content to be naturally fused within the attention mechanism. Compared with dual-stream or cross-attention designs, this single-stream formulation is more streamlined and avoids alignment issues caused by separately processing conditioning and generation signals. The architecture comprises 16 Transformer layers with approximately 2 billion parameters. Each layer uses standard self-attention and FFN blocks, with RMSNorm \cite{zhang2019root} applied before and after each Transformer block. This design stabilizes activation magnitudes, improves training stability, and enhances generalization, particularly when extrapolating to different image resolutions. We employ 3D Rotary Position Embeddings \cite{su2024roformer} (3D RoPE) to encode multi-dimensional positional information. The two spatial dimensions encode the $(x, y)$ coordinates of latent patches in image space, while the temporal dimension encodes the denoising timestep $t$. This 3D RoPE design not only strengthens spatial–temporal position awareness but also improves the model's ability to extrapolate to different resolutions and aspect ratios during inference.
\subsection{AR--Diffusion Feature Alignment Module}
\label{sec:architecture_ardf}

\begin{figure}[H]
\centering
\includegraphics[width=0.95\linewidth]{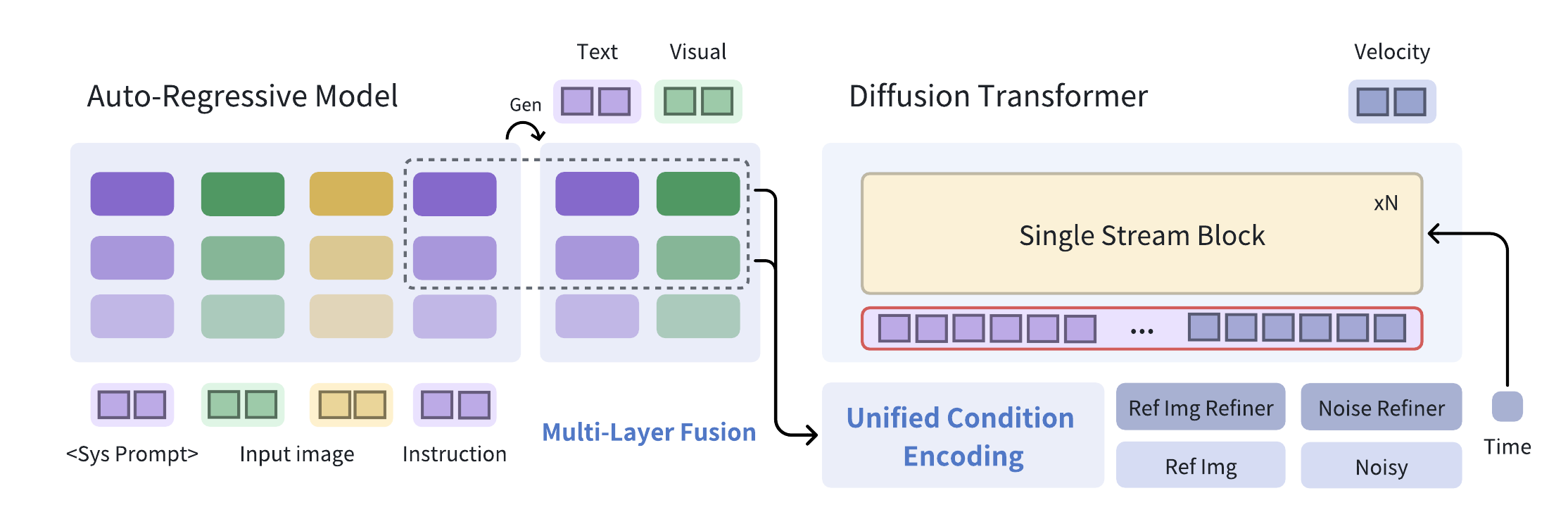}
\caption{\textbf{AR--Diffusion feature alignment module.} Multi-layer feature aggregation, unified condition encoding, and in-context conditioning enable seamless transition from AR feature outputs to diffusion feature inputs.}
\label{fig:feature_alignment}
\end{figure}

One of Mammoth2's core contributions is the efficient coordination between autoregressive reasoning and diffusion generation through a carefully designed feature alignment module. This module integrates Multi-layer Feature Aggregation, Unified Condition Encoding, and In-context Conditioning Injection, forming a synergistic pipeline from ``feature extraction $\rightarrow$ cross-modal fusion $\rightarrow$ condition injection'' that progressively realizes a seamless transition from AR feature outputs to diffusion feature inputs.

\textbf{Multi-layer Feature Aggregation.} In multimodal reasoning tasks, Qwen3-VL processes input text autoregressively and progressively generates discrete visual tokens. During this process, hidden states at different depths encode rich multi-granularity information: early layers preserve fine-grained features from the visual encoder, including spatial structure, texture details, and local patterns; middle layers balance semantic abstraction and visual detail, capturing object-level features and mid-level compositional information; later layers emphasize high-level semantic understanding and global layout planning, with representations closer to the semantic space of language tokens. Traditional methods typically use only the final-layer features of the VLM as conditioning signals, which leads to clear limitations: the last layer is biased toward high-level abstraction and struggles to retain fine-grained visual details and spatial structure, resulting in single-granularity semantics. Moreover, single-layer features are vulnerable to layer-specific representation biases. In contrast, multi-layer feature aggregation provides more stable and comprehensive conditioning signals that jointly account for details and semantics~\cite{li2025unifusionvisionlanguagemodelunified}. We employ average pooling to aggregate multi-layer hidden states, providing rich multi-granularity information for subsequent condition encoding.

\textbf{Unified Condition Encoding.} To effectively fuse text and visual features output by the VLM, Mammoth2 adopts a Unified Condition Encoding mechanism. After multi-layer feature aggregation, text features are projected into the DiT hidden space through a linear layer, while visual features are compressed from variable length to a fixed 64 tokens via an independent Q-Former \cite{li2023blip} (a 2-layer cross-attention module with 64 learnable queries), which shortens the input sequence and reduces computation while preserving key visual information. The preprocessed text and visual features are then concatenated along the sequence dimension and fed into a shared 2-layer bidirectional transformer for unified encoding.

\textbf{In-context Conditioning Injection.} Features processed by the Unified Condition Encoding module are injected into the DiT via in-context conditioning. Concretely, the refined conditioning features are concatenated with VAE-encoded noisy latents along the sequence dimension to form a unified input sequence. The DiT performs full-sequence self-attention over this concatenated sequence, enabling natural interaction and fusion between AR conditioning features and VAE noisy latents within the attention mechanism.

Our feature alignment module progressively aligns the AR and diffusion feature spaces in three stages. Compared with simple MLP-based or Transformer-based alignment interfaces, it offers two key advantages: (1) \textbf{Multi-granularity feature fusion}: through multi-layer feature aggregation, the model simultaneously exploits fine-grained visual details from early layers, object-level features from intermediate layers, and high-level semantic understanding from later layers, providing the diffusion model with comprehensive and stable conditioning signals and avoiding the single-granularity limitation of single-layer features; (2) \textbf{Unified representation learning}: by fusing text and visual features into a shared space via a bidirectional Transformer and enabling early cross-modal interaction, combined with in-context injection, we achieve a seamless transition from AR reasoning to diffusion generation. Details of layer selection strategies, query-token configurations, and ablation results are presented in the Experiments section (Section~\ref{sec:experiments}).

\section{Training Strategy and Data}
\label{sec:data}

In this section, we first describe the multi-stage training recipe used to realize the AR--Diffusion architecture introduced in Section~\ref{sec:method}, and then summarize the data mixtures that support each stage.

\subsection{Training Strategy}

\begin{figure}[H]
    \centering
    \includegraphics[width=\linewidth]{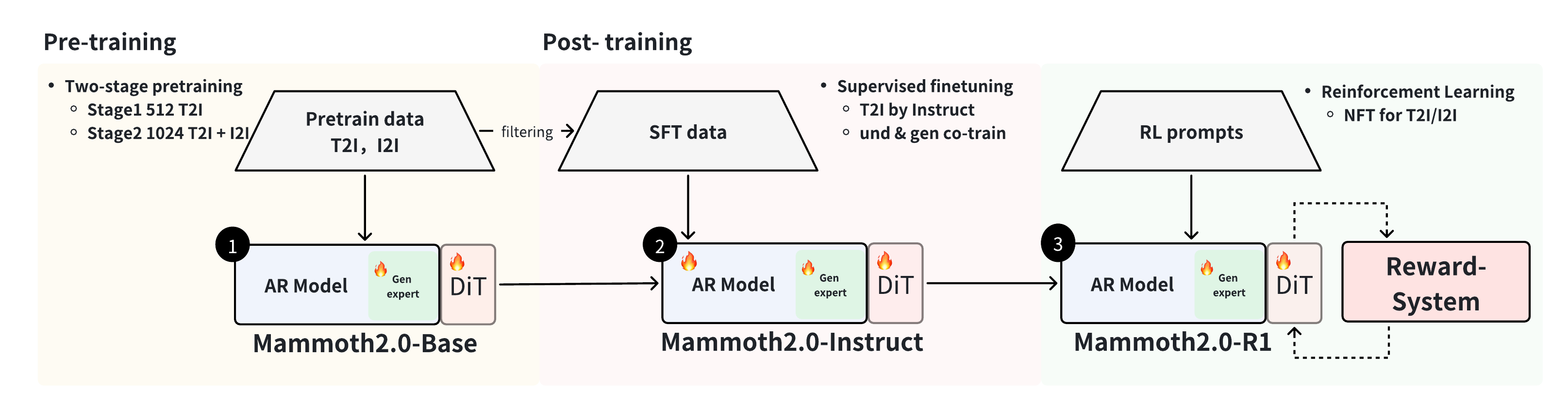}
    \caption{\textbf{Multi-stage training strategy}. Stage 1: generation pretraining; Stage 2: unified joint training with all parameters unfrozen (SFT); Stage 3: RL post-training (DiffusionNFT on the diffusion branch) with multi-signal rewards.}
    \vspace{-2mm}
    \label{fig:training_strategy}
\end{figure}

% Pre-training
\subsubsection{Pre-training}

The pre-training phase aims to introduce general generation and editing capabilities atop the multimodal understanding backbone. To avoid compromising the original backbone's general understanding metrics, we train only the newly added generation-related parameters during this stage, including Gen Expert (initialized from text parameters), Gen Emb/Head, and DiT-related parameters, while freezing all vision-understanding-related parameters. This phase utilizes a large-scale pure generation task dataset, encompassing both text-to-image generation and editing tasks. We therefore jointly optimize the AR next-token predictor and the diffusion velocity predictor with two objectives:

% 1. NTP Loss：仅包含生成
% $$\mathcal{L(U)} = \sum_i\log P(u_i|u_{i-k}, \cdots, u_{i-1};\theta)$$
1. \textbf{Cross-entropy loss} (next-token prediction):
$$\mathcal{L}_{\text{NTP}} = \sum_i\log P(u_i|u_{i-k}, \cdots, u_{i-1};\theta)$$

% 2. Flow Matching Loss
% $$\mathcal{L}_{Flow}(\theta)= \mathbb{E}_{(x_1, Q) \sim \mathcal{D}, t \sim \mathcal{U}(0, 1), X_0 \sim \mathcal{N}(0,1)}\left[||V(X_t, Q, t)-V_t\|^2 \right]$$
2. \textbf{Flow-matching loss} (velocity prediction):
$$\mathcal{L}_{\text{flow}}(\theta)= \mathbb{E}_{(x_1, Q) \sim \mathcal{D}, t \sim \mathcal{U}(0, 1), X_0 \sim \mathcal{N}(0,1)}\left[\|V(X_t, Q, t)-V_t\|^2 \right]$$

We design a progressive training strategy, dividing pre-training into two sub-stages:
\begin{itemize}
\item \textbf{Stage 1}: Training images are constrained to a maximum resolution of 512×512; only text-to-image generation data are trained in this stage. The DiT accepts only text features from the autoregressive backbone as conditioning input. This stage primarily trains generation-related parameters from scratch, establishing initial text-to-image generation capabilities.
\item \textbf{Stage 2}: Image editing data are introduced in this stage, with maximum resolution constrained to 1024×1024. The DiT then fully accepts both text and visual features from the backbone, allowing the model to learn complex instruction-based image editing on top of the stable text-to-image capabilities established in Stage 1.
\end{itemize}

\subsubsection{Post-training}

\textbf{Supervised Fine-Tuning (SFT).} Building upon pre-training, we introduce high-quality understanding and generation training data and unfreeze all model parameters. For the generative part, we start from the large text-to-image and editing pool used in pre-training and apply a two-stage filtering pipeline: (1) rule-based filtering using resolution constraints and aesthetic scoring to discard clearly low-quality samples; and (2) automatic scoring with strong vision-language models to assess semantic alignment, instruction following, and text rendering quality, keeping only high-scoring examples and forming a compact but high-quality SFT generation mixture. Multimodal understanding data are constructed separately, based on open-source instruction-following datasets such as LLaVA-OneVision-1.5 with answers regenerated by Doubao-1.5-Vision-Pro to obtain high-quality supervision. The resulting SFT dataset contains comparable scales of text-to-image samples, editing samples, and general understanding samples. Consistent with pre-training, the DiT accepts text features and image features from the MLLM backbone as conditioning, with an NTP loss training the AR next-token predictor over image and text tokens, and a flow-matching loss training the diffusion velocity predictor for image synthesis. Through multi-task collaborative training, we achieve synchronized improvements in both understanding and generation metrics, obtaining truly unified multimodal capabilities. The key stage-wise training configurations for pre-training and SFT are summarized in Tab.~\ref{tab:train_config_pre_sft}.

\begin{table}[t]
    \centering
    \small
    \caption{Stage-wise training configurations for pre-training (two sub-stages) and supervised fine-tuning (SFT). ``\checkmark'' indicates trainable parameters and ``\ding{55}'' indicates frozen parameters. The corresponding data mixtures are summarized in the Data subsection of this section (Section~\ref{sec:data}).}
    \label{tab:train_config_pre_sft}
    \tablestyle{6pt}{1.2}
    \begin{tabular}{lccc}
        \toprule
        & \textbf{Pre-Stage 1} & \textbf{Pre-Stage 2} & \textbf{SFT} \\
        \midrule
        \multicolumn{4}{l}{\textbf{Trainable modules}} \\
        Backbone (Qwen3-VL-8B)          & \ding{55} & \ding{55} & \checkmark \\
        Gen experts / Emb / Head    & \checkmark & \checkmark & \checkmark \\
        DiT decoder \& alignment module & \checkmark & \checkmark & \checkmark \\
        \midrule
        \multicolumn{4}{l}{\textbf{Optimization}} \\
        LR (Und params)                  & - & - & 1e-5 \\
        LR (Gen experts / Emb / Head)    & 1e-4 & 2e-5 & 1e-5 \\
        LR (DiT decoder)                 & 1e-4 & 2e-5 & 1e-5 \\
        LR scheduler                     & warmup + constant & warmup + constant & warmup + cosine \\
        Max resolution        & $\le 512^2$ & $\le 1024^2$ & $\le 1024^2$ \\
        Max sequence length   & $16{,}384$ & $16{,}384$ & $16{,}384$ \\
        \midrule
        \multicolumn{4}{l}{\textbf{Objectives and data}} \\
        Loss weight (NTP : VP) & 1 : 1 & 1 : 1 & 1 : 1 \\
        Data / tasks  & T2I only & T2I + edit & T2I + edit + Und \\
        Task sampling & batch-level mix & batch-level mix & task-balanced mini-batch mix \\
        \bottomrule
    \end{tabular}
\end{table}

\textbf{Reinforcement Learning (RL)} is introduced to further enhance text-to-image generation and image editing capabilities in a single unified stage. Instead of training separate policies for different tasks, we optimize one diffusion-based policy jointly on both T2I and IT2I samples drawn from the compact RL dataset in Section~\ref{sec:data}. Unlike traditional policy gradient-based RL methods (e.g., GRPO \cite{shao2024deepseekmath}), we adopt DiffusionNFT (Diffusion Negative-aware Fine-Tuning ~\cite{zheng2025diffusionnft}) as our core policy optimization algorithm. This method is naturally compatible with the Flow Matching forward process, supports high-order solvers, and requires no likelihood estimation, thereby enabling more efficient and stable training. We now describe the multi-dimensional reward system that provides optimization signals for this RL stage.

% DiffusionNFT Algorithm：在 Flow Matching 目标上执行策略优化，通过对比学习机制引导模型速度预测器 $$v_θ$$ 向高奖励策略靠近，同时远离低奖励策略。其核心损失函数定义为：
%                        $$\mathcal{L}(\theta) = \mathbb{E}_{c,\pi^{\text{old}}(x_0|c),t}\Big[ r \cdot \|v^+_\theta(x_t,c,t)-v\|_2^2 + (1-r) \cdot \|v^-_\theta(x_t,c,t)-v\|_2^2 \Big]$$
\textbf{DiffusionNFT}: This performs policy optimization on the Flow Matching objective, guiding the model's velocity predictor $v_\theta$ toward high-reward policies while steering away from low-reward policies through a contrastive learning mechanism, which could be formulated as:
\[
\mathcal{L}(\theta) = \mathbb{E}_{c,\pi^{\text{old}}(x_0|c),t}\Big[ r \cdot \|v^+_\theta(x_t,c,t)-v\|_2^2 + (1-r) \cdot \|v^-_\theta(x_t,c,t)-v\|_2^2 \Big]
\]

% 其中 $$r \in [0,1]$$ 为优化概率（由奖励信号归一化得到），$v^+_θ$ 和 $$v^-_θ$$ 分别为隐式正负策略，通过超参数 β 与旧策略 $$v^{\text{old}}$$ 和当前策略 $$v_θ$$ 的线性组合构成：
%                                                $$v^+_\theta = (1-\beta)v^{\text{old}} + \beta v_\theta, \quad v^-_\theta = (1+\beta)v^{\text{old}} - \beta v_\theta$$
where $r \in [0,1]$ is the optimality probability (obtained by normalizing the reward signal), and $v^+_\theta$ and $v^-_\theta$ are implicit positive and negative policies, respectively, constructed as linear combinations of the old policy $v^{\text{old}}$ and current policy $v_\theta$ via hyperparameter $\beta$:
\[
v^+_\theta = (1-\beta)v^{\text{old}} + \beta v_\theta, \quad v^-_\theta = (1+\beta)v^{\text{old}} - \beta v_\theta
\]

% 该设计解耦了训练与采样过程，允许在采样阶段使用任意黑盒求解器（如 DPM-Solver），显著提升采样效率与生成质量。
This design decouples training and sampling processes, allowing the use of arbitrary black-box solvers (e.g., DPM-Solver~\cite{lu2022dpm}) during sampling, significantly improving sampling efficiency and generation quality.

% 我们在原本的 NFT 基础上进行改进：生成与编辑联合训练，针对编辑任务新增 timestep 维度与 spatial 维度的损失加权策略。多任务协同优化：扩展支持生成与编辑任务的联合训练，通过统一奖励体系实现跨任务的指标协同提升。细粒度控制：针对编辑任务特性，时间步维度（timestep weighting） 和 空间维度（spatial weighting） 的损失加权策略，重点优化关键编辑区域和关键去噪阶段，进一步提升局部编辑精度与内容一致性。
We introduce improvements upon the original NFT: joint training of generation and editing, with newly added loss weighting strategies in timestep and spatial dimensions for editing tasks.

% Reward System：
\subsubsection{Reward System}

% 文本到图像生成在实际场景中需同时满足语义对齐、审美质量、人类偏好与文本渲染等多维诉求。单一指标（如 CLIP 或审美分）难以全面刻画，易导致优化偏置。为此，我们混合多维度奖励系统，将多类型、互补的奖励模型加权融合，提供覆盖全面、可解释且可调的优化信号，服务于基于强化学习的对齐与提升，并且在统一阶段进行生成/编辑的训练。系统由多个子奖励组成：HPSv3（人类偏好）、Aesthetic V2（审美质量）、OCR reward、UnifiedReward（MLLM 统一质量评价）、GenEval（结构化评分），Qwen2.5‑VL‑32B多模态打分。其中：
Text-to-image generation in practical scenarios must simultaneously satisfy multiple demands including semantic alignment, aesthetic quality, human preferences, and text rendering. Single metrics (e.g., CLIP \cite{radford2021learning} or aesthetic scores) cannot comprehensively capture these aspects and easily lead to optimization bias. Therefore, we employ a hybrid multi-dimensional reward system that integrates diverse, complementary reward models through weighted fusion, providing comprehensive, interpretable, and adjustable optimization signals for RL-based alignment and improvement, with unified training of generation and editing conducted in a unified stage. The system comprises multiple sub-rewards: HPSv3 \cite{ma2025hpsv3} (human preference), Aesthetic V2 \cite{schuhmann2022aesthetic} (aesthetic quality), OCR reward, UnifiedReward \cite{wang2025unified} (MLLM unified quality assessment), GenEval (structured scoring), and Qwen2.5-VL-32B \cite{bai2025qwen2} multimodal scoring. Specifically:

\textbf{Human Preference Score}: The HPSv3 model effectively predicts human preferences for generated images and exhibits strong generalization across diverse image distributions, serving as a core component for guiding RL toward ``more aesthetically pleasing and preference-aligned" optimization. Compared to HPSv2~\cite{wu2023human}, HPSv3 achieves higher correlation with human preferences (Spearman $\approx$ 0.94 vs 0.87), stronger generalization, and more comprehensive and reliable assessment.

\textbf{OCR Accuracy Score}: Text rendering accuracy is one of the key challenges in text-to-image generation. For prompts requiring text generation within images, we employ OCR-based rewards: comparing rendered text with expected text and quantifying fidelity, jointly evaluating images to compute text rendering accuracy scores. This signal provides critical guidance for improving stability and clarity in text-containing image generation scenarios.

\textbf{Text-Image Alignment Score}: To ensure semantic consistency between prompts and generated images, we leverage VLMs such as Qwen2.5-VL-32B to compute alignment rewards, measuring the correspondence between textual descriptions and visual content, both encouraging contextual relevance and suppressing semantic hallucinations (content unrelated to text) as much as possible.

\textbf{Instruction-based Editing Reward}: For image editing tasks, we further deploy a Qwen2.5-VL-32B judge (invoked via the internal \texttt{bes.general\_audit.qwen25\_vl\_32b} service) that evaluates triples of (source image, edited image, editing instruction) and assigns a discrete score in the range $[0,5]$, reflecting both instruction-following accuracy and visual quality of the edited result. Instead of directly using the greedy-decoded integer as reward, we obtain the full log-probability distribution over candidate numeric scores and take a probability-weighted expectation as the final reward. This expectation-based reward is significantly more stable and reliable than naive greedy decoding, and empirically leads to smoother optimization signals in RL. We apply these reward models on the compact RL dataset described in Section~\ref{sec:data}.

\subsubsection{Mitigating the Train--Test Gap in Joint AR--Diffusion Training}
In our serial AR--Diffusion design, whenever the diffusion decoder is jointly optimized with the autoregressive backbone (namely, in Pre-Stage~2 and the SFT stage in Fig.~\ref{fig:training_strategy}), it is conditioned on discrete visual tokens predicted by the AR path. This makes the overall system particularly vulnerable to teacher-forcing mismatch: during training, the AR path always receives ground-truth tokens, whereas at inference it must condition on its own imperfect predictions, leading to rapid error accumulation across the autoregressive generation. To mitigate this exposure bias, we introduce a lightweight noise regularization applied to the AR input stream during these joint-training phases, perturbing MammothTok visual tokens so the model learns to remain stable under degraded contexts. We investigate two perturbation schemes---\emph{Region Noise}, which corrupts spatially contiguous patches, and \emph{Similarity Noise}, which substitutes tokens with semantically proximate alternatives from the MammothTok codebook. Empirically and through visualization, Similarity Noise proves to be more effective: it injects meaningful variability without disrupting global structure. Thus, we adopt Similarity Noise as our default strategy for alleviating teacher-forcing issues in our AR--DiT architecture whenever AR and diffusion are trained jointly.

\begin{figure}[H]
\centering
\includegraphics[width=\linewidth]{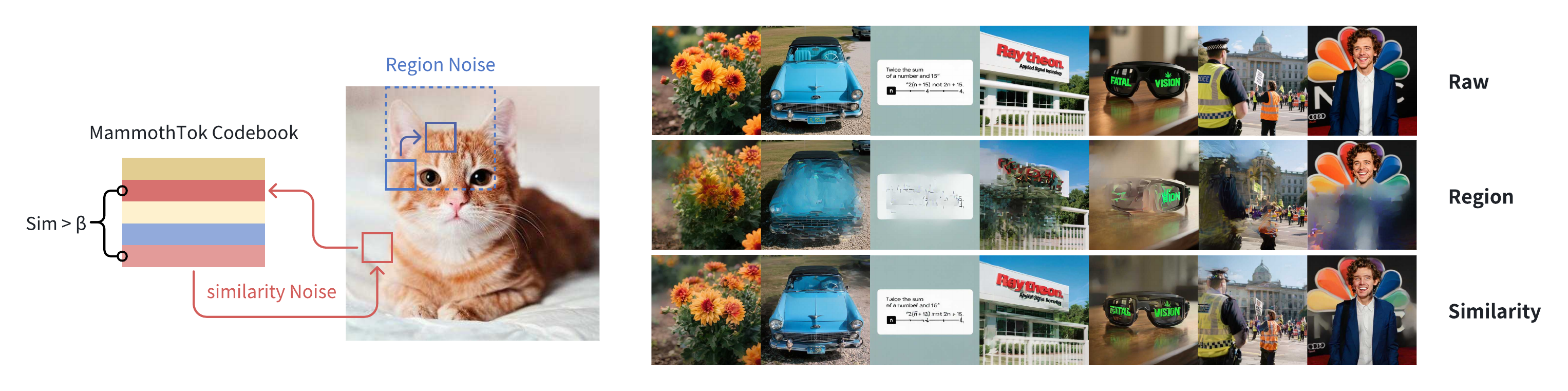}
\caption{\textbf{Noise Regularization on MammothTok Tokens.} We visualize Region Noise (spatial patch corruption) and Similarity Noise (codebook-based token replacement) applied to MammothTok discrete tokens during training. Compared with Region Noise, Similarity Noise better preserves global structure while injecting realistic local variations, leading to more robust autoregressive trajectories under teacher forcing at inference time.}
\vspace{-2mm}
\label{fig:teacher_forcing}
\end{figure}

\subsection{Data}
Having outlined our training pipeline and regularization strategies, we now summarize the data mixtures that feed into each stage. Mammoth2 is trained on a large-scale multimodal corpus built around tens of millions of generation-centric samples for text-to-image (T2I) and image-text-to-image (IT2I) tasks, complemented by additional image-text-to-text (IT2T) and Any-to-Any (A2A) data. In terms of task coverage, the T2I data encompass general generation, text-guided generation, and prompt-based generation, while the IT2I data provide a comprehensive suite of editing capabilities, including general editing, OCR editing, segmentation and extraction, and specialized editing. IT2T data cover multimodal instruction-following tasks such as VQA, captioning, and OCR, and A2A data extend generation and editing with CoT-enhanced reasoning. The dataset includes bilingual (Chinese and English) data with prompts of varying lengths to meet multilingual requirements. For quality assurance, we employ aesthetic scoring and resolution filtering, reverse data pairing, and multiple rounds of iterative refinement, ensuring continuous quality improvement and robust learning. Regarding data sources, we combine the diversity of open-source datasets with large-scale internally constructed data to cover specific scenarios, providing abundant and high-quality training material. A high-level breakdown of the main data categories is deferred to Tab.~\ref{tab:data_overview} in the appendix.

% 4.1 文生图数据（T2I）
\paragraph{\textbf{Text-to-Image (T2I) Data}}

The T2I dataset contains tens of millions of samples and serves as the primary source for training the model's image generation capabilities. It combines large-scale internal synthetic data with high-aesthetic open-source images filtered by aesthetic scores and resolution, and further incorporates specialized subsets (e.g., Flux Reason, BLIP3o) to strengthen performance on complex prompts. Together with the IT2I editing data described below, a more detailed categorization of T2I/IT2I data is summarized in Tab.~\ref{tab:data_overview} in the appendix.

\paragraph{\textbf{Image-Text-to-Image (IT2I) Data}}

The IT2I image editing dataset contains tens of millions of samples, where the model edits an input image conditioned on textual instructions. It covers general editing, text editing, segmentation and extraction, face editing, and pose/action editing with bilingual instructions. The dataset combines diverse open-source editing corpora, large-scale internal datasets with forward–backward pairing, and specialized subsets for fine-grained editing capabilities. All T2I and IT2I subsets are summarized jointly in Tab.~\ref{tab:data_overview}.

\textbf{Open-source data} account for several million samples, providing diverse editing scenarios and high-quality annotations for model training. For general editing, this includes datasets such as NHR-Edit \cite{kuprashevich2025nohumansrequired}, GPT-IMAGE-EDIT-1.5M \cite{wang2025gpt} and ImgEdit, as well as large-scale video editing data from the OmniGen2-X2I2 \cite{wu2025omnigen2} series.

\textbf{Internally constructed data} form the bulk of the dataset and are all derived from open-source raw images. They include large-scale general editing pairs with forward–backward pairing, text editing data in both Chinese and English, segmentation and extraction data, and specialized editing subsets for faces and human pose/action, covering a wide range of real-world editing scenarios.

\paragraph{\textbf{Image-Text-to-Text (IT2T) Data }}
The IT2T data consists of multimodal instruction-following data widely used in existing MLLMs, covering common multimodal tasks such as VQA, captioning, and OCR. Specifically, we utilize open-source datasets from LLaVA-OneVision-1.5 \cite{li2024llava}, selecting a few million samples and regenerating answers using Doubao-1.5-Vision-Pro~\cite{doubao2025}, which significantly reduces computational cost while preserving as much as possible the strong performance of the underlying backbone.

\paragraph{\textbf{Any-to-Any (A2A) Data}}
The A2A data comprises carefully selected generation and editing data, for which we use Doubao-1.5-Vision-Pro to generate CoT annotations and model them as A2A tasks, stimulating the model's understanding and reasoning abilities for complex generative scenarios, without disclosing specific corpus scales.

\paragraph{\textbf{RL Training Data}} We construct a compact RL dataset that covers both text-to-image generation and image editing. For text-to-image, we subsample a small but diverse set of prompts from the broader T2I pool (on the order of a few thousand), with coverage of long-form instructions, compositional attributes, and fine-grained text rendering. For editing, we select a subset of high-quality IT2I samples with diverse edit types and strong automatic quality scores, totaling roughly $10$K examples. Overall, this RL set is much smaller than the full T2I and IT2I corpora introduced above, but sufficiently diverse and well-curated to provide stable and informative reward signals for optimizing both generation and editing.

\subsection{Infrastructure and Acceleration}

% --------------------------------- 版本5 shentao精简 ---------------------------------
Training Mammoth2 requires scaling optimization and infrastructure to more than 10 billion parameters under highly variable sequence lengths. We adopt Fully Sharded Data Parallel (FSDP) with the ``FULL\_SHARD'' strategy to shard model parameters, gradients, and optimizer states across devices, and combine it with activation checkpointing and automatic mixed precision (AMP) in bfloat16 to substantially reduce memory footprint while maintaining stable optimization. To further smooth memory usage, non-zero ranks instantiate the model on the ``meta'' device and materialize weights only after sharded initialization, avoiding construction-time out-of-memory issues.

To improve throughput, we design a lightweight dynamic batching scheme that minimizes both padding waste and GPU idle time. A zero-padding forward mechanism allows us to handle mixed-length sequences without allocating full-length dense tensors, while a token-budget-based batch packer computes token counts on the fly, prefetches candidate samples, and greedily fills each GPU up to a target token budget. This yields a roughly 90\% relative improvement in average GPU Streaming Multiprocessor (SM) utilization and close to $3\times$ end-to-end training throughput, without loss in training accuracy (excluding the DiT module). In addition, we employ FlashAttention-2~\cite{dao2023flashattention}, fused kernels such as fused RMSNorm, and overlapped data prefetching and asynchronous checkpointing to further reduce memory bandwidth overhead and pipeline stalls, resulting in more efficient utilization of large-scale GPU clusters.
\section{Benchmark Evaluations}
\label{sec:eval}

\subsection{Overview}

We first provide a comprehensive comparison of Mammoth2 against existing state-of-the-art models across multiple modalities in Tab.~\ref{tab:compare}. 
This overview establishes Mammoth2's position as a unified multimodal model that achieves competitive or superior performance across understanding, generation, and editing tasks.

\begin{table*}[!htbp]
\centering
\caption{Comparison of different models across multimodal understanding, image generation, and image editing benchmarks. \textsuperscript{\textdagger} refers to methods using an LLM-based prompt rewriter.}
\label{tab:compare}
\tablestyle{6pt}{1.2}
\resizebox{\textwidth}{!}{%
\begin{tabular}{l|c|c|ccc|cc|cc}
\toprule
& & & \multicolumn{3}{c|}{\textbf{Understanding}} & \multicolumn{2}{c|}{\textbf{Image Generation}} & \multicolumn{2}{c}{\textbf{Image Editing}} \\
\textbf{Model} & \textbf{\# Params} & \textbf{Generation Data} & \textbf{MMB} & \textbf{MMMU} & \textbf{MM-Vet} & \textbf{GenEval} & \textbf{DPG} & \textbf{ImgEdit} & \textbf{GEdit-EN} \\

\midrule
\rowcolor{gray!20}\multicolumn{10}{l}{\textit{Unified Models w/ Pre-trained Generator}} \\
UniPic 2.0 & 7B + 2B & 150M & 83.5 & 58.6 & 67.1 & 0.90\textsuperscript{\textdagger} & 83.79 & 4.06 & 7.10 \\
MetaQuery-XL & 7B + 1.6B & 27.4M & 83.5 & 58.6 & 66.6 & 0.80\textsuperscript{\textdagger} & 82.05 & -- & -- \\
UniWorld-V1 & 7B + 12B & 2.7M & 83.5 & 58.6 & 67.1 & 0.84\textsuperscript{\textdagger} & 81.38 & 3.26 & 4.85 \\
X-Omni & 7B + 12B & 600B tokens & 74.8 & -- & -- & 0.83 & 87.65 & -- & -- \\
LightFusion & 7B + 5B + 3B & 45M & 83.5 & 58.6 & 67.1 & 0.91\textsuperscript{\textdagger} & 82.16 & 3.77 & 6.06 \\
\midrule
\rowcolor{gray!20}\multicolumn{10}{l}{\textit{Unified Models w/o Pre-trained Generator}} \\
NExT-OMNI & 7B & 45M & 78.9 & 43.7 & 40.1 & 0.85 & 84.46 & -- & -- \\
Janus-Pro & 7B & 144M & 75.5 & 36.3 & 39.8 & 0.80 & 84.19 & -- & -- \\
Emu3 & 8B & 13T tokens & 58.5 & 31.6 & 37.2 & 0.66\textsuperscript{\textdagger} & 80.60 & -- & -- \\
BLIP3-o 4B & 3B + 1.4B & 28M %218M 
& 78.6 & 46.6 & 60.1 & 0.81\textsuperscript{\textdagger} & 79.36 & -- & -- \\
BLIP3-o 8B & 7B + 1.4B & 61M
%218M 
& 83.5 & 58.6 & 66.6 & 0.84\textsuperscript{\textdagger} & 81.60 & -- & -- \\
Ovis-U1 & 2.4B + 1.2B & -
%6M
& 77.8 & 51.1 & 66.7 & 0.89 & 83.72 & 4.00 & 6.42 \\
Show-o2 & 7B & 82M
%144M 
& 79.3 & 48.9 & -- & 0.76 & 86.14 & -- & -- \\
OmniGen & 3.8B & 100M & -- & -- & -- & 0.68 & 81.16 & 2.96 & 5.06 \\
OmniGen2 & 3B + 4B & 150M & 79.1 & 53.1 & 61.8 & 0.86\textsuperscript{\textdagger} & 83.57 & 3.44 & 6.42 \\
BAGEL & 7B + 7B & 1600M & \textbf{85} & 55.3 & 67.2 & 0.88\textsuperscript{\textdagger} & 85.07 & 3.20 & 6.52 \\
Lumina-DiMOO & 8B & 65M & 84.5 & 58.6 & - & 0.88 & 86.04 & -- & -- \\
OneCAT & 3B & 115M & 78.8 & 41.9 & 52.2 & 0.90 & 84.53 & 3.43 & -- \\
\midrule
Mammoth2 & 8B + 3B + 2B & 60M & 84.2 & \textbf{67.6} & \textbf{73.8} & 0.87 & \textbf{87.20} & \textbf{4.06} & \textbf{6.60} \\
\bottomrule
\end{tabular}
}
\end{table*}

\noindent\textbf{Key Observations.} 
As shown in Tab.~\ref{tab:compare}, Mammoth2 exhibits several key strengths:
(i) \textit{Understanding}: Achieves 84.2\% on MMB and 73.8\% on MM-Vet, delivering strong multimodal comprehension comparable to understanding-only models while surpassing unified baselines such as BAGEL~\citep{deng2025emerging};
(ii) \textit{Generation}: Attains 87\% on GenEval and 87.20\% on DPG, surpassing most unified models and approaching specialized generation systems like SD3-medium (74\%, 84.08\%), despite using substantially fewer supervised generation samples (about 60M vs.\ 1600M for BAGEL);
(iii) \textit{Editing}: Reaches 4.06 on ImgEdit, clearly outperforming unified models such as BAGEL (3.20) and OmniGen2~\citep{zhou2025omnigen2} (3.44), and further exceeding LightFusion \cite{wang2025lightbagel} (3.77 on ImgEdit and 6.06 on GEdit-EN) with higher scores across both benchmarks (4.06 and 6.60, respectively);
(iv) \textit{Efficiency}: Attains these results with a compact 13B parameter budget---smaller than LightFusion's 15B---and with only \textbf{about 60M} generation-centric training samples (vs.\ 1600M for BAGEL and a comparable 45M for LightFusion), without relying on any external pretrained generation backbone;
(v) \textit{Comprehensive multimodal understanding}: On a broad suite of perception and reasoning benchmarks (Tab.~\ref{tab:multimodal-understanding}), Mammoth2 closely matches the strong understanding-only model Qwen3-VL-8B-Instruct (e.g., 73.8\% vs.\ 74.1\% on MM-Vet and 67.6\% vs.\ 66.8\% on MMMU) while consistently outperforming unified baselines such as BAGEL across most tasks, indicating that equipping the model with powerful generation does not sacrifice core visual--language understanding;
(vi) \textit{Understanding-enhanced generation}: On the RealUnify UEG direct task (Tab.~\ref{tab:realunify}), prompting Mammoth2 to first produce Chain-of-Thought (CoT) explanations before image synthesis improves the average score from 31.3\% to 41.0\% (about +31\% relative), allowing it to clearly surpass all prior unified models and demonstrating that our unified AR formulation can effectively leverage high-level reasoning to boost controllable visual generation.
Collectively, these results indicate that our unified AR--Diffusion architecture provides an effective and data-efficient framework for bridging understanding and generation within a single model trained on a high-quality multimodal corpus built around \textbf{60M} supervised generation samples.

\begin{table}[!htbp]
\centering
\caption{Evaluation of text-to-image generation ability on GenEval benchmark. $^*$ indicates the reprompting technique. Model sizes in "A + B" indicate separate understanding (A) and generation (B) parameters; models without "+" share parameters.}
\label{tab:geneval}
\tablestyle{6pt}{1.1}
\resizebox{\textwidth}{!}{%
\begin{tabular}{l c c c c c c c | c}
\toprule
Method & Params & Single Obj.$\uparrow$ & Two Obj.$\uparrow$ & Counting$\uparrow$ & Colors$\uparrow$ & Position$\uparrow$ & Color Attr.$\uparrow$ & Overall$\uparrow$ \\
\midrule
\rowcolor{gray!20} \multicolumn{9}{l}{\textit{Generation Models}} \\
LlamaGen & 0.8B & 0.71 & 0.34 & 0.21 & 0.58 & 0.07 & 0.04 & 0.32 \\
LDM & 1.4B & 0.92 & 0.29 & 0.23 & 0.70 & 0.02 & 0.05 & 0.37 \\
SDv1.5 & 0.9B & 0.97 & 0.38 & 0.35 & 0.76 & 0.04 & 0.06 & 0.43 \\
PixArt-$\alpha$ & 0.6B & 0.98 & 0.50 & 0.44 & 0.80 & 0.08 & 0.07 & 0.48 \\
SDv2.1 & 0.9B & 0.98 & 0.51 & 0.44 & 0.85 & 0.07 & 0.17 & 0.50 \\
DALL-E 2 & 3.5B & 0.94 & 0.66 & 0.49 & 0.77 & 0.10 & 0.19 & 0.52 \\
Emu3-Gen & 8B & 0.98 & 0.71 & 0.34 & 0.81 & 0.17 & 0.21 & 0.54 \\
SDXL & 2.6B & 0.98 & 0.74 & 0.39 & 0.85 & 0.15 & 0.23 & 0.55 \\
DALL-E 3 & - & 0.96 & 0.87 & 0.47 & 0.83 & 0.43 & 0.45 & 0.67 \\
SD3-Medium & 2B & 0.99 & 0.94 & 0.72 & 0.89 & 0.33 & 0.60 & 0.74 \\
Lumina-Image 2.0 & 2.6B & - & 0.87 & 0.67 & - & - & 0.62 & 0.73 \\
HiDream-I1-Full & 17B & 1.00 & 0.98 & 0.79 & 0.91 & 0.60 & 0.72 & 0.83 \\
Seedream 3.0 & - & 0.99 & 0.96 & 0.91 & 0.93 & 0.47 & 0.80 & 0.84 \\
\midrule
\rowcolor{gray!20} \multicolumn{9}{l}{\textit{Unified Models}} \\
Chameleon & 7B & - & - & - & - & - & - & 0.39 \\
LWM & - & 0.93 & 0.41 & 0.46 & 0.79 & 0.09 & 0.15 & 0.47 \\
SEED-X & - & 0.97 & 0.58 & 0.26 & 0.80 & 0.19 & 0.14 & 0.49 \\
Show-o & 1.3B & 0.95 & 0.52 & 0.49 & 0.82 & 0.11 & 0.28 & 0.53 \\
TokenFlow-XL & 14B & 0.95 & 0.60 & 0.41 & 0.81 & 0.16 & 0.24 & 0.55 \\
Janus & 1.5B & 0.97 & 0.68 & 0.30 & 0.84 & 0.46 & 0.42 & 0.61 \\
Janus-Pro-1B & 1.5B & 0.98 & 0.82 & 0.51 & 0.89 & 0.65 & 0.56 & 0.73 \\
Janus-Pro-7B & 7B & 0.99 & 0.89 & 0.59 & 0.90 & 0.79 & 0.66 & 0.80 \\
Ming-Lite-Uni & - & 0.99 & 0.76 & 0.53 & 0.87 & 0.26 & 0.30 & 0.62 \\
Mogao-7B$^*$ & 7B & 1.00 & 0.97 & 0.83 & 0.93 & 0.84 & 0.80 & 0.89 \\
JanusFlow & 1.3B & 0.97 & 0.59 & 0.45 & 0.83 & 0.53 & 0.42 & 0.63 \\
Bagel & 7B + 7B & 0.99 & 0.94 & 0.81 & 0.88 & 0.64 & 0.63 & 0.82 \\
Bagel$^*$ & 7B + 7B & 0.98 & 0.95 & 0.84 & 0.95 & 0.78 & 0.77 & 0.88 \\
OmniGen2 & 3B + 4B & - & - & - & - & - & - & 0.80 \\
Qwen-Image & 7B + 20B & 0.99 & 0.92 & 0.89 & 0.88 & 0.76 & 0.77 & 0.87 \\
Qwen-Image-RL & 7B + 20B & 1.00 & 0.95 & 0.93 & 0.92 & 0.87 & 0.83 & 0.91 \\
\midrule
Mammoth2 & 8B + 3B + 2B & 1.00 & 0.97 & 0.63 & 0.89 & 0.90 & 0.82 & 0.87 \\
\bottomrule
\end{tabular}%
}
\end{table} 

\subsection{Text-to-Image Generation}

We evaluate Mammoth2 on text-to-image (T2I) generation to quantify general generative competence.
Results are reported on two public benchmarks—GenEval and DPGBench—using the raw prompts without prompt rewriting.
GenEval (Tab.~\ref{tab:geneval}) emphasizes object-attribute and compositional compliance; DPGBench (Tab.~\ref{tab:dpg-bench}) comprises 1K dense prompts with fine-grained breakdowns across Global/Entity/Attribute/Relation/Other.
We abbreviate model variants as Mammoth2-AR and Mammoth2 (AR+Diff.).

\textbf{Mammoth2 demonstrates exceptional performance across both benchmarks, establishing new standards for unified multimodal generation.}
On GenEval, Mammoth2 achieves an overall score of \textbf{87\%}, surpassing most comparable-scale unified models and rivaling the performance of specialized generation-only methods like SD3-Medium (74\%).
Notably, Mammoth2 significantly outperforms other understanding-and-generation models: exceeding BAGEL~\citep{deng2025emerging} by \textbf{5 percentage points} (87\% vs.\ 82\%) and TokenFlow-XL~\cite{qu2024tokenflow} by a substantial \textbf{32 percentage points} (87\% vs.\ 55\%).
The model exhibits particularly strong capabilities in compositional understanding, achieving 97\% on Two Objects and 90\% on Position tasks, demonstrating superior spatial reasoning and multi-object coordination.

On DPGBench, Mammoth2 attains an impressive overall score of \textbf{87.20\%}, ranking among the top-tier models and substantially exceeding most unified approaches.
The model shows balanced excellence across all evaluation dimensions: Global (81.16\%), Entity (92.99\%), Attribute (90.16\%), Relation (94.35\%), and Other (84.80\%).
Particularly noteworthy is the outstanding performance in Relation understanding (94.35\%), where Mammoth2 matches or surpasses even specialized generation models, highlighting the effectiveness of our hybrid AR-Diffusion architecture in capturing complex inter-object relationships.
This comprehensive performance validates that Mammoth2 successfully bridges the gap between unified multimodal models and specialized text-to-image generators.

\begin{table}[!htbp]
\centering
\caption{Results on the DPGBench (DPG) benchmark.}
\label{tab:dpg-bench}
\tablestyle{6pt}{1.2}
\begin{tabular}{l c c c c c c c}
\toprule
 Method & Params & Global & Entity & Attribute & Relation & Other & Overall$\uparrow$ \\
\midrule
\rowcolor{gray!20} \multicolumn{8}{l}{\textit{Generation Models}} \\
 SDv1.5 & 0.9B & 74.63 & 74.23 & 75.39 & 73.49 & 67.81 & 63.18 \\
PixArt-$\alpha$ & 0.6B & 74.97 & 79.32 & 78.60 & 82.57 & 76.96 & 71.11 \\
Lumina-Next & 1.7B & 82.82 & 88.65 & 86.44 & 80.53 & 81.82 & 74.63 \\
SDXL & 2.6B & 83.27 & 82.43 & 80.91 & 86.76 & 80.41 & 74.65 \\
Playground v2.5 & - & 83.06 & 82.59 & 81.20 & 84.08 & 83.50 & 75.47 \\
Hunyuan-DiT & 1.5B & 84.59 & 80.59 & 88.01 & 74.36 & 86.41 & 78.87 \\
PixArt-$\Sigma$ & 0.6B & 86.89 & 82.89 & 88.94 & 86.59 & 87.68 & 80.54 \\
Emu3-Gen & 8B & 85.21 & 86.68 & 86.84 & 90.22 & 83.15 & 80.60 \\
DALL-E 3 & - & 90.97 & 89.61 & 88.39 & 90.58 & 89.83 & 83.50 \\
FLUX.1 [Dev] & 12B & 74.35 & 90.00 & 88.96 & 90.87 & 88.33 & 83.84 \\
SD3-Medium & 2B & 87.90 & 91.01 & 88.83 & 80.70 & 88.68 & 84.08 \\
HiDream-I1-Full & 17B & 76.44 & 90.22 & 89.48 & 93.74 & 91.83 & 85.89 \\
Lumina-Image 2.0 & 3.6B & - & 91.97 & 90.20 & 94.85 & - & 87.20 \\
Seedream 3.0 & - & 94.31 & 92.65 & 91.36 & 92.78 & 88.24 & 88.27 \\
\midrule
\rowcolor{gray!20} \multicolumn{8}{l}{\textit{Unified Models}} \\
Emu3 & 8B & - & - & - & - & - & 80.60 \\
TokenFlow-XL & 14B & - & - & - & - & - & 73.38 \\
Janus & 1.5B & 82.33 & 87.38 & 87.70 & 85.46 & 86.41 & 79.68 \\
Janus-Pro-1B & 1.5B & 87.58 & 88.63 & 88.17 & 88.98 & 88.30 & 82.63 \\
Janus-Pro-7B & 7B & 86.90 & 88.90 & 89.40 & 89.32 & 89.48 & 84.19 \\
Mogao-7B & 7B & 82.37 & 90.03 & 88.26 & 93.18 & 85.40 & 84.33 \\
OmniGen2 & 3B + 4B & - & - & - & - & - & 83.57 \\
Qwen-Image & 7B + 20B & 91.32 & 91.56 & 92.02 & 94.31 & 92.73 & 88.32 \\
\midrule
Mammoth2-AR & 8B + 3B & 82.07 & 92.13 & 89.98 & 93.46 & 82.00 & 84.08 \\
Mammoth2 & 8B + 3B + 2B & 81.16 & 92.99 & 90.16 & 94.35 & 84.80 & 87.20 \\
\bottomrule
\end{tabular}
\end{table} 

\subsection{Image Editing}

We further assess instruction-based image editing capabilities on ImgEdit benchmark (Tab.~\ref{tab:imgedit}).
These evaluations demonstrate Mammoth2's versatility in precise image manipulation while maintaining visual quality and semantic coherence.

\begin{table}[!htbp]
\centering
\caption{Comparison on the ImgEdit benchmark. Higher is better.}
\label{tab:imgedit}
\tablestyle{4pt}{1.2}
\begin{tabular}{l l c c c c c c c c c c}
\toprule
Model & Model Size & Action & Add & Adjust & Extract & Replace & Remove & Background & Style & Hybrid & Overall$\uparrow$ \\
\midrule
OmniGen & 3.8B & 3.38 & 3.47 & 3.04 & 1.71 & 2.94 & 2.43 & 3.21 & 4.19 & 2.24 & 2.96 \\
StepIX-Edit & - & 2.52 & 3.88 & 3.14 & 1.76 & 3.40 & 2.41 & 3.16 & 4.63 & 2.64 & 3.06 \\
Bagel & 14B & 4.17 & 3.56 & 3.31 & 1.70 & 3.30 & 2.62 & 3.24 & 4.49 & 2.38 & 3.20 \\
UniWorld-v1 & 19B & 2.74 & 3.82 & 3.64 & 2.27 & 3.47 & 3.24 & 2.99 & 4.21 & 2.96 & 3.26 \\
OmniGen2 & 7B & 4.68 & 3.57 & 3.06 & 1.77 & 3.74 & 3.20 & 3.57 & 4.81 & 2.52 & 3.44 \\
FLUX.1 Kontext [Pro] & 12B & 4.63 & 4.25 & 4.15 & 2.35 & 4.56 & 3.57 & 4.26 & 4.57 & 3.68 & 4.00 \\
GPT Image 1 [High] & - & 4.89 & 4.61 & 4.33 & 2.90 & 4.35 & 3.66 & 4.57 & 4.93 & 3.96 & 4.20 \\
Qwen-Image & 7B + 20B & 4.69 & 4.38 & 4.16 & 3.43 & 4.66 & 4.14 & 4.38 & 4.81 & 3.82 & 4.27 \\
\midrule
Mammoth2 & 8B + 3B + 2B & 4.37 & 4.57 & 4.05 & 3.38 & 4.18 & 3.34 & 4.13 & 4.59 & 3.96 & 4.06 \\
\bottomrule
\end{tabular}
\end{table}

\textbf{Mammoth2 establishes itself as a leading unified model for image editing, achieving performance comparable to specialized editing systems.}
On ImgEdit, Mammoth2 attains an impressive overall score of \textbf{4.06}, substantially outperforming other unified models, including BAGEL~\citep{deng2025emerging} (3.20, +27\% improvement) and OmniGen2~\citep{zhou2025omnigen2} (3.44, +18\% improvement).
This performance closely approaches proprietary state-of-the-art systems like GPT Image 1 [High] (4.20) and Qwen-Image (4.27), demonstrating that our unified architecture does not compromise on specialized editing capabilities.

The model exhibits particularly strong performance in creative editing tasks: achieving 4.57 on Add operations and 4.59 on Style modifications, indicating superior capability in content generation and artistic transformation.
Mammoth2 also demonstrates robust performance across diverse editing scenarios, with consistent scores above 4.0 for Action (4.37), Adjust (4.05), and Hybrid (3.96) tasks.

These results validate that Mammoth2 successfully extends beyond pure generation to deliver comprehensive image manipulation capabilities, positioning it as a versatile solution for diverse visual content creation and modification tasks.

\textbf{Visualization:} Fig.~\ref{fig:main_showcase} showcases qualitative examples on DPGBench, demonstrating the strong instruction-following capability of our model when handling complex prompts with diverse visual properties.

\subsection{Multimodal Understanding}

We evaluate Mammoth2 on a diverse suite of multimodal understanding benchmarks to assess whether our unified generation training compromises core perception capabilities. As shown in the overall comparison (Tab.~\ref{tab:compare}), Mammoth2 maintains robust performance on key understanding metrics, achieving 84.2\% on MMBench, 67.6\% on MMMU, and 73.8\% on MM-Vet. These results indicate that despite being trained with extensive generative objectives, the model effectively retains the strong visual--language comprehension of its Qwen3-VL backbone, showing minimal degradation compared to understanding-only baselines.

In a more detailed comparison (Tab.~\ref{tab:multimodal-understanding}), we benchmark Mammoth2 against the competitive unified model BAGEL~\citep{deng2025emerging} as well as specialized Qwen-VL variants. Benefiting from the powerful Qwen3-VL base model, Mammoth2 consistently outperforms BAGEL across multiple reasoning-intensive tasks such as MMMU and MathVista. Although we observe a slight performance gap relative to the pure understanding-only Qwen3-VL-8B model—likely due to the limited scale of our high-quality understanding dataset during post-training—Mammoth2 strikes a superior balance between understanding and generation compared to prior unified models.

\begin{table}[!htbp]
\centering
\caption{Results on Multimodal Understanding Benchmarks.}
\label{tab:multimodal-understanding}
\tablestyle{4pt}{1.2}
\begin{tabular}{l c c c c c c c c c c }
\toprule
Model & MMVet & OCRBench & MMMU & Hallusion & AI2D & MMStar & MathVista & MMBench & MME & MMVP \\
\midrule
\rowcolor{gray!20} \multicolumn{11}{l}{\textit{Understanding-Only Models}} \\
Qwen2.5-VL-7B-Instruct & 67.1 & 884 & 58.6 & 51.9 & 84.6 & 64.5 & 68.2 & 83.5 & 2347 & 77.0 \\
Qwen3-VL-8B-Instruct & 74.1 & 912 & 66.8 & 59.2 & 85.5 & 71.7 & 77.0 & 84.7 & 2405 & 79.3 \\
\midrule
\rowcolor{gray!20} \multicolumn{11}{l}{\textit{Unified Models}} \\
Bagel & 67.2 & -- & 55.3 & -- & -- & -- & 73.1 & 85.0 & 2388 & 69.3 \\
Mammoth2 & 73.8 & 901 & 67.6 & 58.5 & 84 & 70.7 & 73.6 & 84.2 & 2401 & 79.0 \\
\bottomrule
\end{tabular}
\end{table}

\label{sec:understanding_evals}

\subsection{Understanding Enhances Generation}

\begin{figure}[H]
    \centering
    \includegraphics[width=\linewidth]{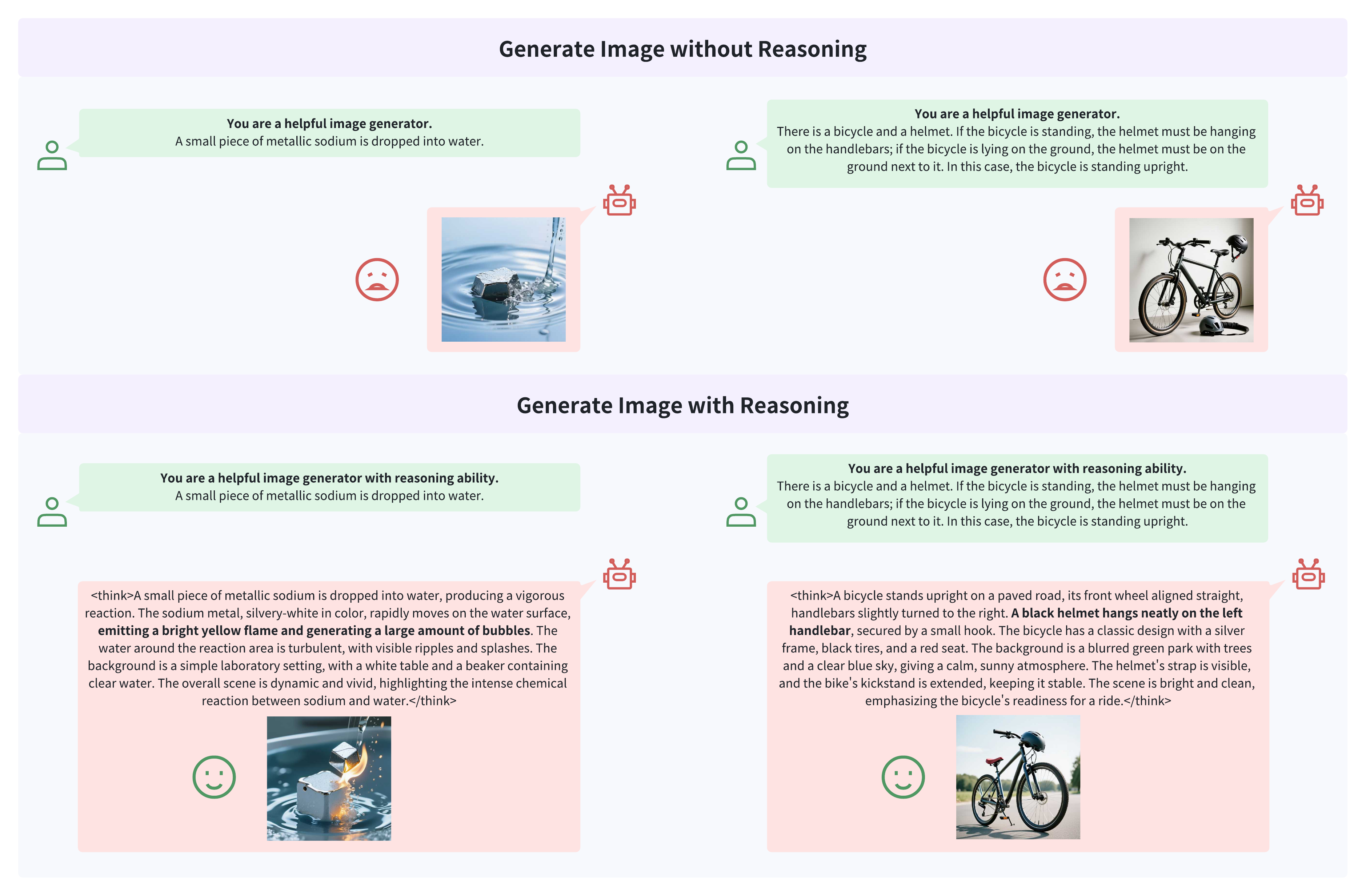}
    \caption{Qualitative examples demonstrating how Chain-of-Thought (CoT) reasoning enhances image generation quality in the UEG benchmark. When Mammoth2 first articulates its reasoning process before generating images, it produces more accurate, contextually grounded, and semantically coherent visual outputs across diverse reasoning categories including world knowledge, commonsense reasoning, mathematical reasoning, and scientific reasoning.}
    \vspace{-2mm}
    \label{fig:show_case}
\end{figure}

We further evaluate Mammoth2 on the RealUnify~\citep{shi2025realunifyunifiedmodelstruly} benchmark, concentrating on the UEG direct task. The results in Tab.~\ref{tab:realunify} provide compelling evidence of the efficacy of our unified autoregressive (AR) paradigm for seamlessly integrating multimodal understanding and generation. Strikingly, when the model is prompted to articulate a Chain-of-Thought (CoT) prior to producing the final image, performance improves from 31.3\% to 41.0\%---a gain of nearly ten points (about +31\% relative) over direct generation. Moreover, this CoT-enhanced decoding strategy enables Mammoth2 to decisively outperform all competing models. Qualitative examples in Fig.~\ref{fig:show_case} illustrate how CoT reasoning helps the model produce more accurate, contextually grounded, and semantically coherent visual outputs across diverse reasoning categories. These findings underscore the robustness of our approach and highlight the benefits of autoregressive unification in strengthening high-level reasoning and controllable visual synthesis.

\begin{table}[!htbp]
\centering
\caption{Evaluation results on RealUnify UEG direct task.WR: World Knowledge; CR: Commonsense Reasoning; MR-I: Mathematical Reasoning; LR: Logical Reasoning; SR: Scientific Reasoning; C2T: Codeto-Image}
\label{tab:realunify}
\tablestyle{8pt}{1.2}
\begin{tabular}{l c c c c c c c}
\toprule
Model & Avg & WK & CR & MR-I & LR & SR & C2I \\
\midrule
BLIP3-o            & 33.0 & 57 & 71 & 21 & 19 & 28 & 2  \\
Show-o2            & 28.0 & 30 & 56 & 25 & 21 & 18 & 18 \\
OminiGen2          & 30.3 & 36 & 61 & 21 & 29 & 16 & 19 \\
BAGEL              & 32.7 & 46 & 70 & 23 & 29 & 21 & 7  \\
\midrule
Mammoth2 w/o CoT           & 31.3 & 56 & 38 & 14 & 26 & 25 & 29 \\
\rowcolor{gray!12}
Mammoth2 w/ CoT       & \textbf{41.0(+9.7)} & 80 & 51 & 20 & 27 & 33 & 35 \\
\bottomrule
\end{tabular}
\end{table}

\section{Experiments}
\label{sec:experiments}

Unless otherwise stated, ablation studies are conducted using a prototype model setup (e.g., smaller training steps or earlier checkpoints) to save computational resources. Consequently, absolute metric scores in this section may differ from the final results of the fully trained Mammoth2 model reported in Section~\ref{sec:eval}.

\subsection{\textbf{AR--Diffusion Design Choices}}
\label{subsec:ar_diffusion_design_choices}
We present comprehensive ablation studies to isolate and quantify the contributions of our key architectural and design choices.

\paragraph{\textbf{Generation Expert Placement}}\label{para:expert_structure}
We study how many layers should specialize as generation experts and find that only a moderate range of deeper layers is necessary. Using a Qwen2.5-VL prototype with identical backbone and training settings (Tab.~\ref{tab:gen_expert_ablation}), the baseline with \emph{FFN+Attention experts at all layers} achieves an average score of 74.05. For \emph{FFN-only experts}, 7 layers is insufficient (AVG: 69.18), 21 layers provides moderate gains (AVG: 76.22), while 14 layers yields the best performance (AVG: \textbf{77.15}). These results show that specializing a mid-to-late window of layers optimally balances semantic understanding and visual generation, synergizing with our AR--Diffusion feature alignment—which primarily consumes mid-to-late AR features—while also reducing parameter and compute overhead. We observe consistent trends on the full Mammoth2 system built on Qwen3-VL-8B, further supporting our expert-decoupling design.

\begin{table}[!htbp]
\centering
\caption{Ablation on placement of generation experts. We report the overall average (AVG) and benchmark scores on GenEval and DPGBench. Higher is better.}
\label{tab:gen_expert_ablation}
\tablestyle{8pt}{1.2}
\begin{tabular}{lccc}
\toprule
Configuration & GenEval & DPGBench & AVG \\
\midrule
FFN+Attention experts all layers & 66.48 & 81.62 & 74.05 \\
FFN experts 21-layers & 72.38 & 80.06 & 76.22 \\
FFN experts 14-layers & 75.48 & 78.81 & 77.15 \\
FFN experts 7-layers & 63.96 & 74.40 & 69.18 \\
\bottomrule
\end{tabular}
\end{table}

\begin{table}[!htbp]
\centering
\caption{Ablation on visual conditional encoder. We report the benchmark scores on ImgEdit. Higher is better.}
\label{tab:gen_ar_visual_condition}
\tablestyle{8pt}{1.2}
\begin{tabular}{c|ccc}
\toprule
Conditions & ViT & MammothTok & All \\
\midrule
ImgEdit & 2.88 & 2.95 & 3.44 \\
\bottomrule
\end{tabular}
\end{table}

\begin{figure}[H]
    \centering
    \includegraphics[width=\linewidth]{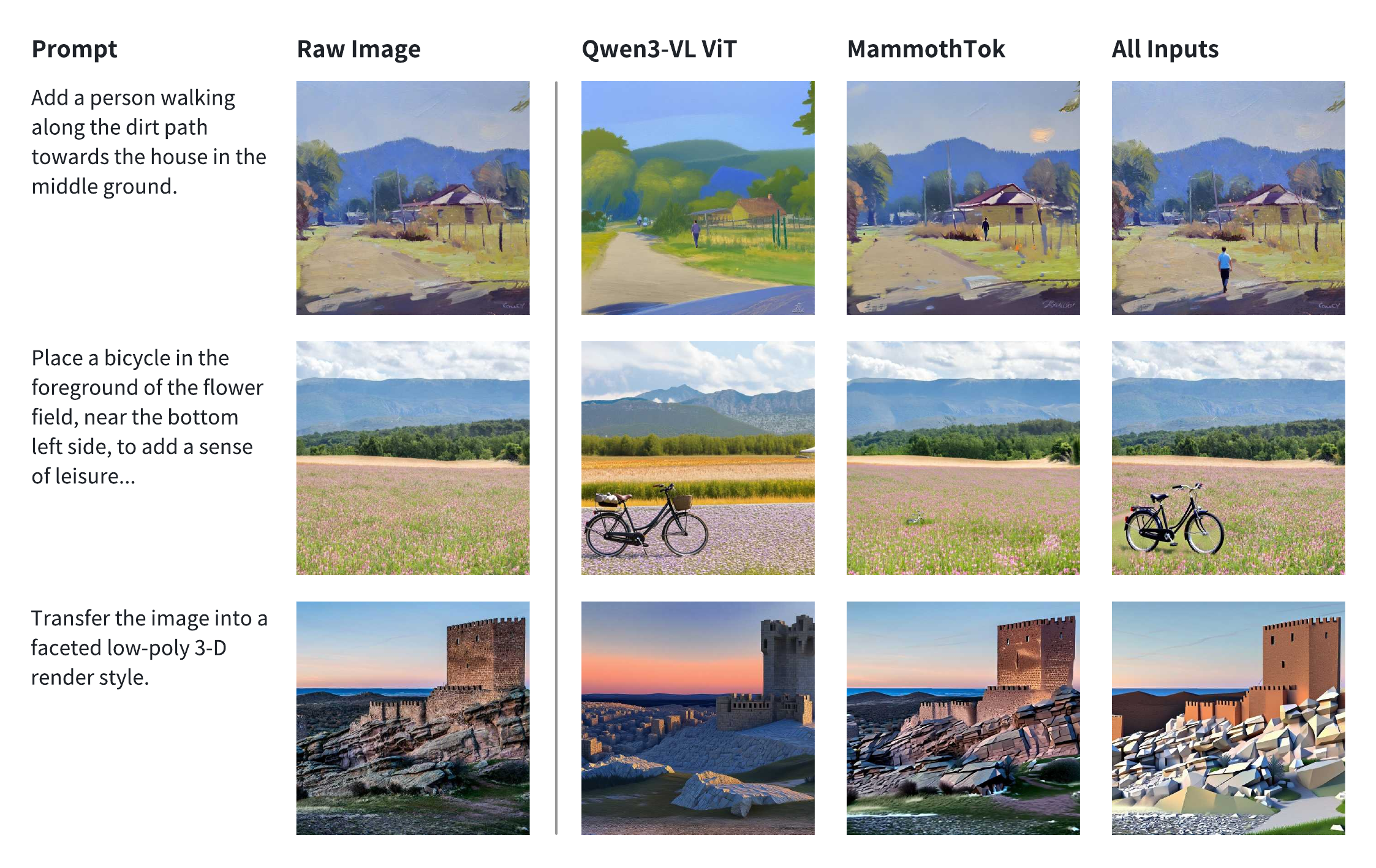}
    \caption{Qualitative ablation study on visual conditioning for instruction-based image editing. Different conditioning strategies (Qwen-only, MammothTok-only, and Qwen+MammothTok) demonstrate varying trade-offs between semantic control, structural preservation, and visual fidelity. The full model combining both visual sources achieves the best overall balance.}
    \vspace{-2mm}
    \label{fig:edit_ablation}
\end{figure}

\paragraph{\textbf{Task-specific Visual Conditioning Encoders}}
We ablate the choice of visual conditioning for \emph{instruction-based image editing}. In addition to the default Qwen3-VL continuous encoder, Mammoth2 also incorporates MammothTok discrete visual tokens (Fig.~\ref{fig:feature_alignment}). Under identical training settings, we compare three variants: using only Qwen features, only MammothTok features, or their combination.
Qwen-only conditioning yields strong semantic alignment but loses fine-grained detail, while MammothTok-only conditioning preserves local appearance yet drifts toward reconstruction with weaker instruction following. Combining both sources achieves the best trade-off between semantic control, structural consistency, and visual fidelity, and is adopted as our default configuration (Tab.~\ref{tab:gen_ar_visual_condition}, Fig.~\ref{fig:edit_ablation}).

\paragraph{\textbf{Multi-layer AR Feature Aggregation}}
Our AR--Diffusion feature alignment module aggregates AR features from multiple Transformer layers before feeding them into the diffusion decoder. To understand which layers are most beneficial, we compare several variants: using only the final AR layer, using only the penultimate AR layer, and uniformly averaging a set of mid-to-late AR layers. As show in Tab.~\ref{tab:arch_multi_layer_fusion}, we find that relying solely on the last layer, while strong, tends to over-emphasize high-level semantics and slightly degrade fine-grained geometry and texture, whereas using only the penultimate layer under-utilizes global context. In contrast, aggregating a small window of mid-to-late layers consistently yields better GenEval and DPGBench scores. These results confirm that \emph{multi-layer feature aggregation} is important for effectively coupling the AR backbone with the diffusion decoder, allowing the latter to simultaneously leverage low-level structure and high-level semantics.

\begin{table}[!htbp]
\centering
\caption{Ablation on Multi-layer AR feature choice. We report the overall average (AVG) and benchmark scores on GenEval and DPGBench. Higher is better}
\label{tab:arch_multi_layer_fusion}
\tablestyle{6pt}{1.2}
\begin{tabular}{lccc}
\toprule
Configuration & GenEval & DPGBench & AVG \\
\midrule
final layer & 71.71 & 85.52 & 78.62 \\
penultimate layer & 71.04 & 85.78 & 78.41 \\
A set of mid-to-late layers & \textbf{72.82} & \textbf{85.91} & \textbf{79.37} \\
\bottomrule
\end{tabular}
\end{table}

\paragraph{\textbf{Coupling Strategy: AR vs AR--Diffusion}}\label{para:arch_ar_diff}
This ablation provides the main empirical test of our architectural hypothesis: that coupling autoregressive and diffusion pathways yields complementary benefits beyond either approach alone. We keep all training settings fixed and only vary (i) whether the diffusion branch is enabled and (ii) whether the AR pathway generates visual tokens at inference. Concretely, we compare three variants:
\begin{itemize}
    \item \textbf{AR}: Pure autoregressive baseline without any diffusion branch.
    \item \textbf{AR+Diff (w/o AR Gen)}: AR serves purely as a feature encoder (for text or text+image inputs), and the diffusion module produces the final pixel outputs and edits; no discrete visual tokens are generated by AR.
    \item \textbf{AR+Diff (w/ AR Gen)}: Joint generation mode where AR first predicts visual tokens encoding structure and layout, which are then refined into high-fidelity pixels by the diffusion decoder.
\end{itemize}

We evaluate all variants on GenEval, DPGBench, and ImgEdit (higher scores indicate better performance; ImgEdit is not evaluated under the w/ AR Gen protocol). Using \emph{Mammoth2-AR} as the reference (GenEval: 81.35, DPGBench: 84.08, ImgEdit: 3.32), enabling the diffusion branch under the \textbf{w/o AR Gen} protocol yields small changes on GenEval (81.10, $-0.25$) but clear gains on DPGBench (85.61, $+1.53$) and ImgEdit (4.06, $+0.74$). The \textbf{w/ AR Gen} configuration achieves the strongest overall performance: GenEval improves to 82.03 ($+0.68$) and DPGBench to 87.12 ($+3.04$), with particularly large gains on DPGBench. These results support our design intuition: the AR pathway is well-suited for semantic planning and language--vision alignment, while the diffusion decoder specializes in fine-grained detail synthesis and editing flexibility; their synergistic combination delivers the largest benefits, especially on compositional generation tasks. Detailed results are presented in Tab.~\ref{tab:arch_ar_diff}.

\begin{table}[!htbp]
\centering
\caption{Architecture comparison of AR baseline versus AR+Diffusion. Definitions of ``w/o AR Gen'' and ``w/ AR Gen'' follow the setup in Paragraph~\ref{para:arch_ar_diff}. Bold indicates the best per column.}
\label{tab:arch_ar_diff}
\tablestyle{6pt}{1.2}
\begin{tabular}{lccc}
\toprule
Configuration & GenEval & DPGBench & ImgEdit \\
\midrule
AR & 81.35 & 84.08 & 3.32 \\
AR+Diff (w/o AR Gen) & 81.10 (-0.25) & 85.61 (+1.53) & \textbf{4.06} (+0.74) \\
AR+Diff (w/ AR Gen) & \textbf{82.03} (+0.68) & \textbf{87.12} (+3.04) & -- \\
\bottomrule
\end{tabular}
\end{table}

\subsection{Reinforcement Learning}
\label{subsec:rl_results}

Building on the SFT checkpoint, we further perform a Negative-aware Fine-Tuning (NFT) stage to jointly enhance both text-to-image generation and image editing. Instead of relying on conventional policy-gradient RL, we directly optimize the Flow Matching velocity predictor $v_\theta$: for each condition $c$, we sample multiple candidate trajectories from the old policy, score them with a multi-dimensional reward model, and normalize these rewards into an optimality probability $r\in[0,1]$ that weights the positive and negative velocity targets in the DiffusionNFT loss (Section~\ref{sec:data}). During this stage, we keep most autoregressive backbone parameters frozen and only fine-tune generation-related components (including selected generation experts and the diffusion decoder).

For reward design, we follow the multi-dimensional Reward System introduced in Section~\ref{sec:data}, combining HPSv3 human preference scores, aesthetic scores, OCR-based text rendering accuracy, UnifiedReward-based holistic quality scores, Qwen2.5-VL-32B-based text–image alignment scores, and the instruction-based editing reward into a unified optimality signal via normalization and clipping. For T2I samples from the compact RL dataset in Section~\ref{sec:data}, only the generation-related rewards are active, while for IT2I samples we additionally upweight OCR, alignment, and the editing reward to emphasize text fidelity and local edit consistency. To systematically study reward configurations, we train NFT variants with single rewards (e.g., HPS-only, OCR-only) as well as a hybrid-reward NFT model where multiple rewards are linearly combined. In practice, we use a learning rate of $1\times10^{-6}$, around 200 optimization steps, a global batch size of 256, and 16 rollouts per prompt. A KL regularization term with weight $0.001$ is added on top of the NFT loss to constrain policy deviations, striking a balance between training stability and exploration.

\begin{figure}[H]
    \centering
    \includegraphics[width=\linewidth]{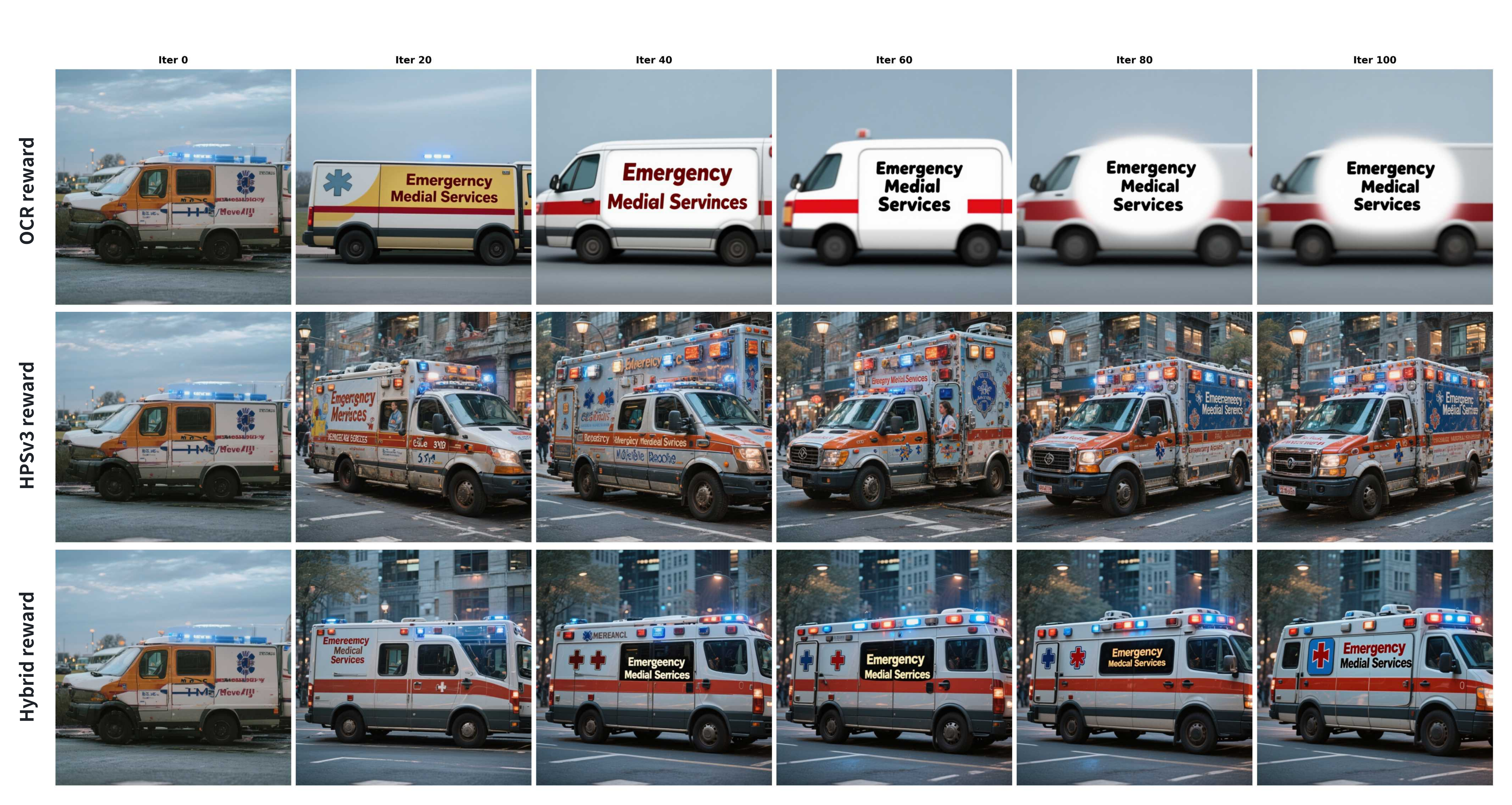}
    \caption{Single-reward training exhibits pronounced reward hacking (see step 100 for OCR and HPSv3). Hybrid reward mitigates this issue, producing images with high visual quality and accurate text rendering.}
    \vspace{-2mm}
    \label{fig:reward_hacking}
\end{figure}

% Reward visualizations
\begin{figure}[H]
    \centering
    \includegraphics[width=\linewidth]{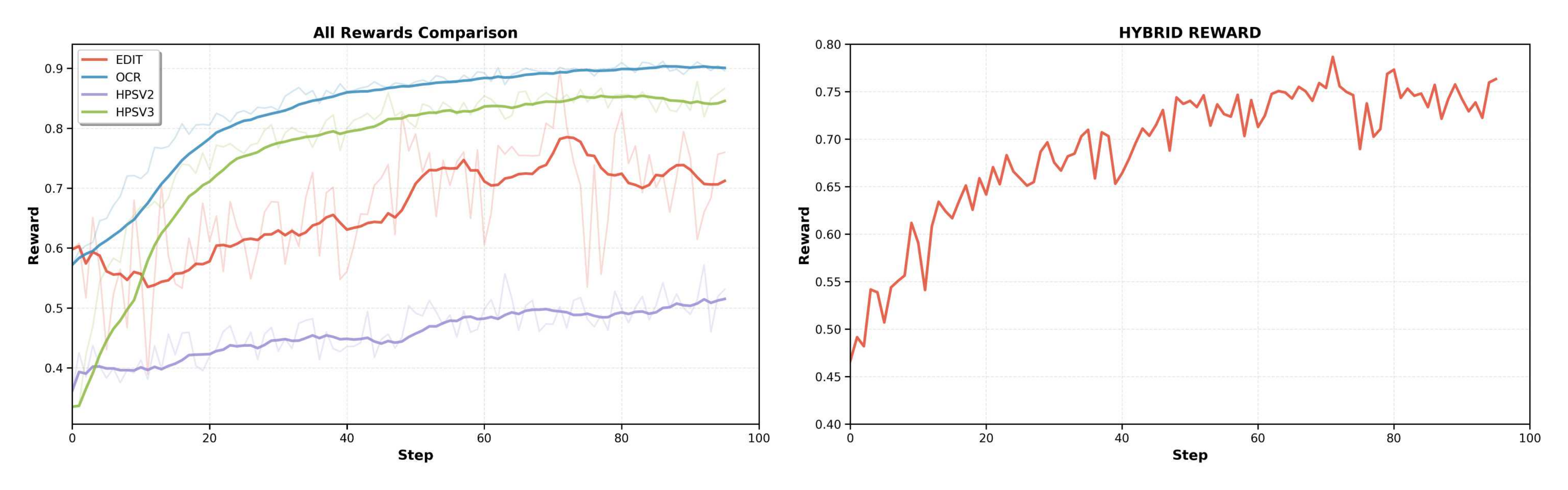}
    \caption{Training curves: single-reward vs.\ hybrid-reward NFT.}
    \vspace{-2mm}
    \label{fig:reward_curve}
\end{figure}

\paragraph{\textbf{Finding 1: NFT rapidly improves w/o-CFG performance}}
As shown in Fig.~\ref{fig:reward_curve}, across all reward configurations (HPS-only, OCR-only, and the hybrid setting), the corresponding reward scores rise quickly and largely saturate within roughly 100 NFT steps, indicating fast convergence of the diffusion policy even when evaluated without classifier-free guidance. Qualitatively, Fig.~\ref{fig:reward_hacking} further illustrates that the w/o-CFG baseline at step 0 produces low-quality, poorly aligned images, whereas after NFT training the same prompts yield substantially sharper, more semantically aligned generations. This behavior is consistent with the fast-improvement regime reported in the DiffusionNFT paper~\cite{zheng2025diffusionnft}.

\paragraph{\textbf{Finding 2: Hybrid reward mitigates reward hacking}}
As illustrated in Fig.~\ref{fig:reward_hacking}, using a single reward such as OCR or HPSv3 leads to clear signs of reward hacking. With OCR-reward-only training, the model rapidly improves text legibility in the early iterations (Iter 20–40), but by Iter 80–100 it starts over-optimizing the text region: the ambulance panel is unrealistically enlarged and the overall composition becomes distorted, even though the rendered words are readable. In contrast, the HPS-reward-only setting drives the model to produce visually appealing images, but also encourages HPSv3-style reward hacking, often yielding overly intricate, detail-saturated renderings. The hybrid-reward NFT model, however, converges at a similar speed while avoiding these failures: across all iterations, the ambulance geometry and scene layout remain realistic, and the key phrases (e.g., “Emergency Medical Services”) stay clear, correctly spelled, and naturally integrated into the scene. This confirms that hybrid rewards effectively mitigate reward hacking, delivering both high visual quality and accurate text rendering.

\section{Conclusion}
\label{sec:conclusion}

We presented MammothModa2 (Mammoth2), a unified AR--Diffusion framework that integrates multimodal understanding, text-to-image generation, and instruction-based image editing within a single model. By coupling an autoregressive backbone with a diffusion decoder through an in-context conditioning interface and the MammothTok visual tokenizer, Mammoth2 cleanly separates semantic planning from high-fidelity pixel synthesis while operating in a shared discrete token space. Generation experts on the AR path, a single-stream DiT decoder, and a carefully designed AR--Diffusion feature alignment module jointly enable strong compositional control and high-quality image synthesis under complex, long-context instructions. Although our implementation is instantiated on Qwen3-VL-8B, the framework itself is compatible with other strong multimodal backbones and can be ported to future VLM architectures.

Our multi-stage training strategy further consolidates these capabilities. Pre-training establishes generation and editing on top of a strong vision-language backbone using progressive curricula and joint NTP + Flow Matching objectives; post-training introduces multimodal supervision and DiffusionNFT-based reinforcement learning over both generation and editing. Across benchmarks, Mammoth2 delivers competitive or superior performance compared with existing unified models: it achieves strong text-to-image scores on GenEval and DPGBench, state-of-the-art performance among unified models on ImgEdit, and preserves robust multimodal understanding competitive with understanding-only backbones such as Qwen3-VL-8B. Notably, these results are obtained with a relatively modest training corpus and without relying on pre-trained generative models, highlighting Mammoth2's parameter and data efficiency.

Looking ahead, we identify several promising directions to advance unified multimodal intelligence:
\begin{itemize}[leftmargin=1.5em, itemsep=0.25em, topsep=0.25em]
    \item \textbf{Unified Video Tokenization and Generation.} We plan to develop next-generation visual tokenizers with higher compression rates that natively support spatiotemporal data, extending Mammoth2's AR--Diffusion framework to unified video understanding and generation.
    \item \textbf{Scaling Laws for Hybrid Architectures.} We aim to systematically investigate the scaling properties of AR--Diffusion models, specifically examining how performance scales with mixture-of-experts routing strategies, DiT parameterization, and training data size to guide efficient scale-up.
    \item \textbf{Unified RL for Perception and Generation.} We will explore unified reinforcement learning paradigms that jointly optimize autoregressive planning and diffusion synthesis, aiming to align both multimodal comprehension and creative generation with human preferences in a single loop.
\end{itemize}

\newpage
\section{Contributors}

\noindent \textbf{Algorithm}: Tao Shen*, Xin Wan*, Taicai Chen, Rui Zhang, Junwen Pan, Dawei Lu

\vspace{0.3em}
\noindent \textbf{Infra}: Fanding Lei, Zhilin Lu, Yunfei Yang, Chen Cheng

\noindent \textbf{Project Leaders}: Qi She, Chang Liu\textsuperscript{$\dagger$}, Zhenbang Sun

\blfootnote{* These authors contributed equally to this work.}
\blfootnote{$\dagger$ Project leader.}

\clearpage
\bibliographystyle{assets/plainnat}
\bibliography{main}

\clearpage
\appendix
% Reset table counter and use ``A.1, A.2, ...''-style numbering for appendix tables
\setcounter{table}{0}
\renewcommand{\thetable}{\thesection.\arabic{table}}

\section{MammothTok}

\subsection{Model Architecture}
\label{subsec:arch}

\begin{figure}[H]
\centering
\includegraphics[width=0.9\linewidth]{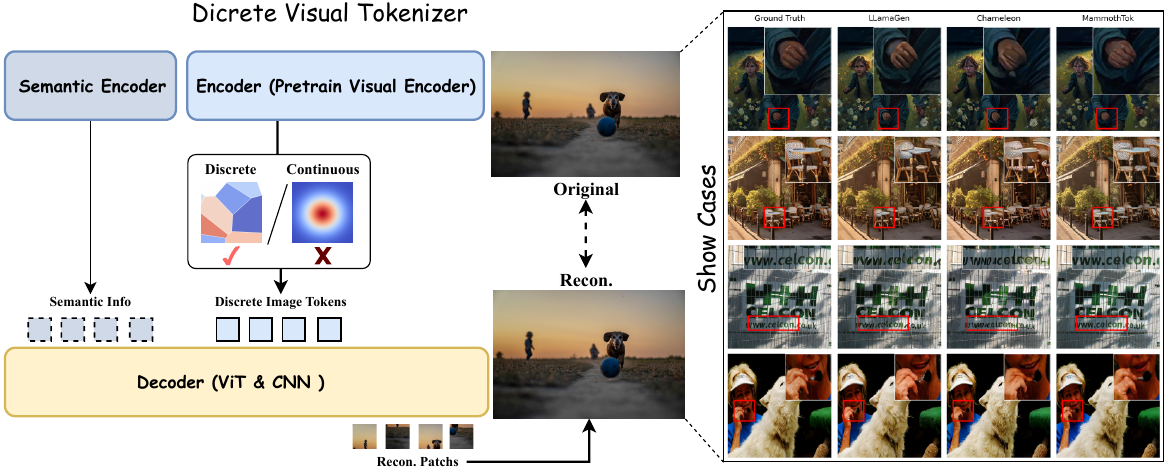}
\caption{\centering \textbf{MammothTok Architecture and Examples.} }
\vspace{-2mm}
\label{fig:mammoth_arch}
\end{figure}

As illustrated in Fig.~\ref{fig:mammoth_arch} (left), MammothTok is built upon the VQ-GAN framework, employing a pretrained visual encoder as the encoder and adopting a hybrid ViT--CNN architecture as the decoder $f^{dec}$. The CNN component captures and reconstructs local details, while the ViT component facilitates scaling up the model capacity. To enable arbitrary-resolution input processing, we replace the original absolute positional encoding of ViT with 2D RoPE, leveraging its extrapolation capability. Previous work has shown that high-level semantic information can improve the reconstruction ability of tokenizers. To this end, we introduce a semantic encoder $f^{enc}_{sem}$ and incorporate a semantic alignment loss~\citep{xiong2025gigatok} to inject richer high-level semantics into the visual tokens, thereby enhancing MammothTok's performance. 
% \paragraph{Semantic Loss} To infuse semantic information into MammothTok's visual tokens, we introduce a semantic alignment loss following the setup of GigaTok~\citep{xiong2025gigatok}. 
Formally, let $f^{dec,l}(x)$ denote the features from the $l$-th Transformer layer of the decoder, and $f^{enc}_{sem}(x)$ denote the semantic features from a pretrained semantic encoder. The loss is defined as:
\begin{equation}
    \mathcal{L}_{sem} = \frac{1}{N}\sum_{n=1}^N sim\big(\phi{(f^{dec,l}(x_n))}, f^{enc}_{sem}(x_n)\big),
\end{equation}
where $N$ is the batch size, $sim(\cdot,\cdot)$ is cosine similarity, and $\phi(\cdot)$ is an MLP projection head that maps $f^{dec}$ to match the dimensionality of $f^{enc}_{sem}$.
As illustrated in Fig.~\ref{fig:mammoth_arch} (right), MammothTok exhibits superior capability in reconstructing fine-grained image details compared with LLamaGen and Chameleon, underscoring the effectiveness of its overall design. Furthermore, with the integration of OCR data during subsequent training, MammothTok achieves substantial improvements in text reconstruction, further validating its design choices.

\subsection{Data and Training Strategy}
\label{subsec:strategy}
MammothTok is trained on large-scale data, augmented with synthetic and OCR data to enhance downstream performance. A total of 100M samples are used, with a 3:1 ratio of natural to synthetic data, and OCR samples comprising one-fifth of the natural data. To reduce training cost, we adopt a curriculum-based resolution training strategy: (1) low-resolution pretraining for basic reconstruction capability; (2) resolution expansion fine-tuning with progressively increased resolution; and (3) native-resolution fine-tuning using original image resolutions. This progressive training scheme achieves a favorable trade-off between performance and efficiency, ultimately yielding a tokenizer that supports arbitrary-resolution inputs and exhibits strong robustness to OCR data.

% \begin{figure}[H]
% \centering
% \includegraphics[width=0.8\linewidth]{figures/tokenizer/MammothTok_Train.pdf}
% \caption{\textbf{MammothTok Training Strategy.} }
% \vspace{-2mm}
% \label{fig:mammoth_train}
% \end{figure}

\subsection{Experiments and Results}

We employ the pretrained native-resolution version of the AIMv2~\citep{aim} visual encoder as the encoder, and adopt a decoder consisting of 24 Transformer blocks ($d=1024$, 16 heads). The codebook size in Mammoth is fixed at 16,384 with an entry dimension of 8. The semantic encoder is an AIMv2-huge model trained with vision-language alignment, and the alignment is performed at the 3rd decoder layer ($l=3$). Optimization uses AdamW ($\beta_1=0.9$, $\beta_2=0.95$) with weight decay 0.1. The learning rate is warmed up for 5{,}000 steps to $1\times10^{-4}$, followed by cosine annealing to $1\times10^{-5}$. Training is performed on 128 GPUs. Progressive training runs for 10, 1, and 1 epochs, respectively, with initialization from the previous stage. During resolution fine-tuning, the initial learning rate is $2\times10^{-5}$. 
%Loss weights are fixed as $\lambda_{sem}=0.5$. %$\lambda_{rec}=\lambda_{percp}=1.0$, $\lambda_{gan}=0.5$, and $\beta=0.25$. 
\textbf{Result:} We evaluate MammothTok on general datasets (ImageNet-Val), synthetic datasets (MJHK-30K), OCR datasets (ICDAR13), comparing it against widely used discrete visual tokenizers in unified MLLMs. As shown in Tab.~\ref{tab:tok_rec}, MammothTok achieves an rFID of 0.50 on ImageNet-Val with a downsampling ratio of 16, outperforming LLamaGen (0.69) and Chameleon (1.03), while also obtaining higher PSNR (23.10) and SSIM (0.73). 
%On MJHK-30K, MammothTok achieves a lower rFID of 0.90 compared with 1.18 and 1.70 for LLamaGen and Chameleon, respectively, and yields the best PSNR (24.58) and SSIM (0.82), indicating better reconstruction performance in complex synthetic scenarios. 
In particular, on the OCR dataset ICDAR, MammothTok attains the highest PSNR (28.03) and SSIM (0.87), markedly outperforming other tokenizers in visual fidelity and text preservation, despite the rFID being influenced by the structural properties of text. Overall, these results demonstrate that MammothTok consistently improves performance across general and synthetic datasets, while showing strong adaptability to text-sensitive OCR tasks. %underscoring its effectiveness and potential in unified multimodal large language models. 
In addition, we conducted preliminary downstream validation experiments by adopting Qwen3-VL-8B as the LLM backbone and constructing AR-based text-to-image models with either the LLamaGen tokenizer or MammothTok. The models were trained on 20M text–image pairs with an image resolution of 384 for 20K iterations. As shown in Tab.~\ref{tab:tok_gen}, the AR model built with MammothTok achieves 61.78 \% on GenEval and 81.45\% on DPGBench, outperforming the LLamaGen tokenizer by +3.53\% and +1.68\%, respectively. These improvements indicate that MammothTok not only enhances general generation ability
%, as measured by GenEval, 
but also provides more robust semantic consistency and alignment.
%on DPGBench. %underscoring its effectiveness as a tokenizer for autoregressive text-to-image generation.

\begin{table*}[t]
\centering
\begin{minipage}{0.48\linewidth}
\centering
\centering
\captionof{table}{Comparison of rFID, PSNR, and SSIM across different datasets.}
\label{tab:tok_rec}
\tablestyle{6pt}{1.1}
\resizebox{\linewidth}{!}{%
\begin{tabular}{lccccccccc}
\toprule
\multirow{2}{*}{Method} & \multicolumn{3}{c}{ImageNet-Val} & \multicolumn{3}{c}{MJHK-30K} & \multicolumn{3}{c}{ICDAR13} \\
\cmidrule(lr){2-4} \cmidrule(lr){5-7} \cmidrule(lr){8-10}
& rFID$\downarrow$ & PSNR$\uparrow$ & SSIM$\uparrow$ & rFID$\downarrow$ & PSNR$\uparrow$ & SSIM$\uparrow$ & rFID$\downarrow$ & PSNR$\uparrow$ & SSIM$\uparrow$ \\
\midrule
LLamaGen & 0.69 & 22.10 & 0.71 & 1.18 & 23.33 & 0.80 & 22.26 & 25.62 & 0.85 \\
Chameleon    & 1.03 & 21.81 & 0.70 & 1.70 & 22.82 & 0.77 & 24.03 & 25.55 & 0.84 \\
MammothTok   & 0.50 & 23.10 & 0.73 & 0.90 & 24.58 & 0.82 & 14.96 & 28.03 & 0.87 \\
\bottomrule
\end{tabular}}

\end{minipage}
\hfill
\begin{minipage}{0.48\linewidth}
\centering
\centering
\captionof{table}{Performance comparison of different tokenizers on GenEval and DPGBench benchmarks.}
\label{tab:tok_gen}
\tablestyle{6pt}{1.1}
\resizebox{0.75\linewidth}{!}{%
\begin{tabular}{lcc}
\toprule
Tokenizer & GenEval & DPGBench \\
\midrule
LLamaGen   & 58.25\% & 79.77\% \\
MammothTok & 61.78\% (+3.53\%) & 81.45\% (+1.68\%) \\
\bottomrule
\end{tabular}}

\end{minipage}
\end{table*}

\section{Additional Data Details}

\begin{table}[!ht]
    \centering
    \small
    \caption{Overview of MammothModa2 training data. Type codes: T2I (text-to-image), IT2I (image-text-to-image editing), IT2T (image-text-to-text), and A2A (any-to-any).}
    \label{tab:data_overview}
    \tablestyle{4pt}{1.15}
    \begin{tabular}{l >{\raggedright\arraybackslash}p{0.40\linewidth} c c c}
        \toprule
        \textbf{Type} & \textbf{Subset} & \textbf{\#Samples} & \textbf{Lang.} & \textbf{Used in stages} \\
        \midrule
        \multirow{4}{*}{T2I} &
        Internal synthetic &
        $\sim$24M &
        Zh/En &
        \multirow{4}{*}{Pre-1, Pre-2, SFT} \\
        &
        High-aesthetic LAION (filtered) &
        $\sim$8.5M &
        Multi &
        \\
        &
        Flux Reason subset &
        $\sim$2M &
        Multi &
        \\
        &
        BLIP3o series subset &
        $\sim$3.7M &
        Multi &
        \\
        \midrule
        \multirow{8}{*}{IT2I} &
        Open-source general editing (NHR-Edit, GPT-IMAGE-EDIT-1.5M, ImgEdit) &
        $\sim$5.1M &
        Zh/En &
        \multirow{8}{*}{Pre-2, SFT} \\
        &
        OmniGen2-X2I2 video editing &
        $\sim$2M &
        Zh/En &
        \\
        &
        Internal general editing &
        $\sim$12M &
        Zh/En &
        \\
        &
        Internal text editing (Zh/En) &
        $\sim$15M &
        Zh/En &
        \\
        &
        Segmentation \& extraction &
        $\sim$2.8M &
        Zh/En &
        \\
        \midrule
        IT2T &
        LLaVA-OV-1.5 subset &
        $\sim$2M &
        Zh/En &
        SFT \\
        \midrule
        A2A &
        generation & editing tasks w/ CoT &
        Zh/En &
        SFT \\
        \bottomrule
    \end{tabular}
\end{table}

\end{document}